\pdfoutput=1

\documentclass[11pt]{article}

\usepackage[]{acl_style/acl}

\usepackage{times}
\usepackage{latexsym}

\usepackage[T1]{fontenc}
\usepackage[utf8]{inputenc}

\usepackage{microtype}

\usepackage{inconsolata}

\usepackage{amsmath,amsfonts,bm}

\def\eqref#1{equation~\ref{#1}}

\def\1{\bm{1}}

\DeclareMathAlphabet{\mathsfit}{\encodingdefault}{\sfdefault}{m}{sl}
\SetMathAlphabet{\mathsfit}{bold}{\encodingdefault}{\sfdefault}{bx}{n}

\DeclareMathOperator*{\argmax}{arg\,max}
\DeclareMathOperator*{\argmin}{arg\,min}

\usepackage{hyperref}
\usepackage{url}
\usepackage{graphicx}
\usepackage{amsmath,amsfonts,amssymb}
\usepackage{booktabs}
\usepackage{mathtools}
\usepackage{newtxmath} %
\usepackage{enumitem}

\usepackage{chngcntr}
\newcommand{\LM}{{\rm LLM}}

\newcommand{\Think}{\textbf{Think}}
\newcommand{\Sum}{\textbf{Sum}}
\newcommand{\ThinkSum}{\Think\Sum}

\newcommand{\taskname}[1]{\textsc{#1}}

\newcommand{\sumtechnique}[1]{\textbf{\color{black}#1}}
\newcommand{\thinktechnique}[1]{\textbf{\color{black}#1}}

\makeatletter

\def\paragraph{\@startsection{paragraph}{4}{\z@}{0.5ex plus
   0.2ex minus .2ex}{-1em}{\normalsize\bf}}

\makeatother

\title{ThinkSum: Probabilistic reasoning over sets using large language models}

\author{
  Batu Ozturkler \\
  Stanford University \\
  Stanford, California, USA \\
  \texttt{ozt@stanford.edu} \\
  \And
  Esmeralda S. Whitammer\\
  Mila, Universit\'e de Montr\'eal \\
  Montr\'eal, Qu\'ebec, Canada \\
  \texttt{esmeralda.whitammer@mila.quebec} \\
  \AND
  Zhen Wang \\
  Ohio State University \\
  Columbus, Ohio, USA \\
  \texttt{wang.9215@osu.edu} \\
  \And
  Nebojsa Jojic \\
  Microsoft Research \\
  Redmond, Washington, USA \\
  \texttt{jojic@microsoft.com} \\
  }

\begin{document}
\maketitle

\let\tssize\relax
\maketitle
\let\tssize\small

\begin{abstract}
Large language models (LLMs) have a substantial capacity for high-level analogical reasoning: reproducing patterns in linear text that occur in their training data (zero-shot evaluation) or in the provided context (few-shot in-context learning). However, recent studies show that even the more advanced LLMs fail in scenarios that require reasoning over multiple objects or facts and making sequences of logical deductions. We propose a two-stage probabilistic inference paradigm, \ThinkSum, which reasons over sets of objects or facts in a structured manner. In the first stage (\Think{} -- retrieval of associations), a LLM is queried in parallel over a set of phrases extracted from the prompt or an auxiliary model call. In the second stage (\Sum{} -- probabilistic inference or reasoning), the results of these queries are aggregated to make the final prediction. We demonstrate the possibilities and advantages of \ThinkSum{} on the BIG-bench suite of LLM evaluation tasks, achieving improvements over the state of the art using GPT-family models on thirteen difficult tasks, often with far smaller model variants. We also compare and contrast \ThinkSum{} with other proposed modifications to direct prompting of LLMs, such as variants of chain-of-thought prompting. Our results suggest that because the probabilistic inference in \ThinkSum{} is performed outside of calls to the LLM, \ThinkSum{} is less sensitive to prompt design, yields more interpretable predictions, and can be flexibly combined with latent variable models to extract structured knowledge from LLMs. Overall, our proposed paradigm represents a promising approach for enhancing the reasoning capabilities of LLMs.
\end{abstract}

\section{Introduction}

Large language models \cite[LLMs;][]{gpt3,rae2021scaling,chowdhery2022palm} can recall a broad range of basic facts, recognize and mimic various forms in language, and efficiently extrapolate analogies in structure and meaning. These abilities allow LLMs to excel in zero-shot and few-shot tasks formulated as the generation or selection of a likely completion to a prompt. This formulation requires LLMs to perform \textbf{fast associative thinking}, in which each token of text in the sequence making up the answer is generated or scored in one pass through the model and, other than that, no intermediate information is created or retained. This fast thinking is made possible by the compression of information that is repeated in a variety of ways in large training datasets, within the LLM's weights.

However, it is increasingly evident that when \textbf{reasoning}, or slow thinking, is required, failure modes of LLMs are revealed. In our usage, reasoning refers to the sequential manipulation of concepts that can be expressed in language. Tasks that require iterative retrieval of rarely stated knowledge, uncertainties over multiple objects or facts, or multiple steps of deduction are difficult even for the most advanced LLMs~\cite{suzgun2022challenging}. In a recently designed suite of evaluations, BIG-bench \citep{srivastava2022beyond}, some of the tasks where the gap between machine and human performance is large involve inference sequences with nested counterfactuals (\taskname{Logical deduction}), concepts introduced through definitions (\taskname{Conceptual combinations}), etc. (see Fig.~\ref{fig:margins}). These are tasks where a human solver's intuitive feeling of `(in)coherence' is insufficient to produce the right answer, and a sequence of thoughts, along with the use of intermediate results, may be necessary to arrive at the solution, particularly when working memory is insufficient.

\begin{figure*}[t]
    \centering
    \includegraphics[width=\textwidth]{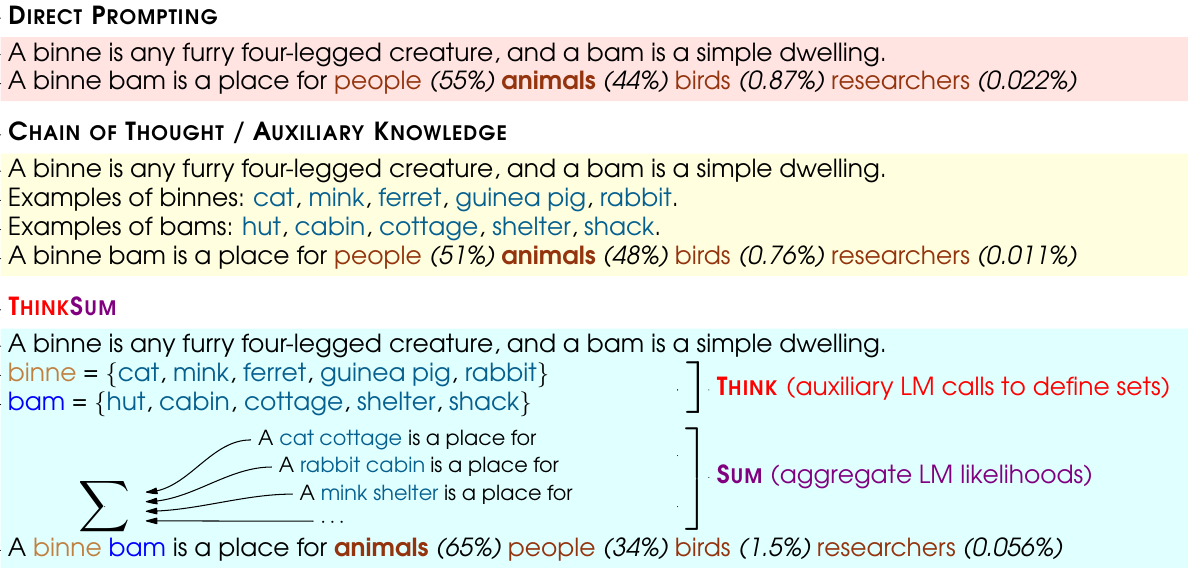}
    \caption{An example adapted from the \taskname{Conceptual combinations (Invented words)} task, in which models must select the most likely completion of a phrase that includes nonce words whose definitions are given. \textbf{Top:} \textbf{Direct prompting} evaluates completion likelihoods normalized over the four answer choices (`people', `animals', `birds', `researchers'). \textbf{Middle:}  \textbf{Chain-of-thought}-like or \textbf{auxiliary knowledge} approaches would query a LLM or knowledge base for additional context. This example shows the brittleness entrusting all `reasoning' to self-attention in linear text, especially in smaller models, which have stronger recency bias \citep{whitammer-etal-2022-coherence}: if we simply list generated examples as the additional context in the prompt, the recency bias causes the LLM to still give a higher probability to `people' than to `animals', simply because `bam' (simple dwelling) examples are given after the `binne' examples. \textbf{Bottom:}  Our \ThinkSum{} approach to this task queries a LLM (GPT-2 XL) to produce sets of examples defining the nonce words, then marginalizes over substitutions of these examples into the target phrase.}
    \label{fig:thinksum_teaser}
\end{figure*}

We show several tasks in BIG-bench that can be addressed by a two-component mechanism, which we name \ThinkSum{}\footnote{\ThinkSum{} is named by analogy with other algorithms with `expand' and `aggregate' steps, such as MapReduce in distributed computing and sum-product in graphical models.}:
\begin{enumerate}[left=0pt,nosep,label=$\bullet$]
    \item \Think{} (fast thinking / association / knowledge retrieval step): creating an association of text spans with sets of strings. This process may involve generation from a language model, as is the case in Fig.~\ref{fig:thinksum_teaser}, where the novel word `binne' is associated with the set of strings $\{\text{`cat'},\text{`mink'},\dots\}$ by prompting GPT-3 with the definition and asking for examples. Alternatively, it may consist solely of a scoring mechanism, resulting in the formation of a matrix of probabilities on which probabilistic inference is performed.
    \item \Sum{} (slow thinking / \Sum{marization} / reasoning step): probabilistic inference that aggregates generated strings or probabilities to produce the final answer. Summarization typically involves, and often entirely consists of, summing of probabilities of strings (computed in the \Think{} step), as in Fig.~\ref{fig:thinksum_teaser}, where the final word is assumed to be sampled from a mixture of possible substitutions of `binne' and `bam' words into the input. 
\end{enumerate}
We discuss different ways to \Think{} and to \Sum{} in section \S\ref{sec:thinksum}, but we start with one example, illustrated in Fig.~\ref{fig:thinksum_teaser} (bottom), motivated by the \taskname{Conceptual combinations (Invented words)} task in BIG-bench. In this task, the LLM is provided with the definitions of two invented words and asked to infer the most plausible sentence that uses a combination of the invented words. As the words are not common or consistently used in the training set, the LLM needs to understand and combine the definitions of the invented words to reason about the meaning of the combination. The LLM is queried to produce example instances of the invented words with the help of the definitions. These example instances can be substituted into the query in place of the invented words. By mapping individual spans of the text of interest to sets, we arrive at a mixture model (in this example, a mixture with 25 components for 5 possible replacements of each word), which can be used in the same manner the original LLM is used, either to score text or to generate it token by token. When we score all candidate completions using this mixture model and normalize over the four choices, the correct  answer -- that `binne bams' are for animals and not people -- becomes the most likely.

An important difference between our \ThinkSum{} and existing chain-of-thought-like prompt engineering methods \citep{wei2022chain,kojima2022large}, is that our reasoning step is not reduced to a generation problem for the LLM, but is performed as a probabilistic inference external to the LLM. This reduces vulnerability to features of the prompt, such as accidental distraction of the LLM by spurious patterns (see Fig.~\ref{fig:thinksum_teaser}, middle). Instead, we engineer the slow thinking process to make parallel calls to the LLM to query for intermediate information, then possibly perform programmatic recombination of strings (\Think). The final reasoning step -- in which likelihoods obtained from the LLM for the recombinations derived from earlier steps of the reasoning process are combined to make the final prediction -- is left to classical probabilistic reasoning (\Sum). In a sense, \Sum{} replaces the self-attention mechanism over linear text, which is used as the sole `reasoning' mechanism in chain-of-thought-like approaches that expect the intermediate `thoughts' to take the form of generated tokens intervening between the input and output.

Imposing an alternative reasoning system over an associative ``knee-jerk reaction" system has an analogy with models of human cognitive processes \citep{tversky1974judgment,kahneman} that separate System 1 (fast thinking) and System 2 (slow thinking). System 2 acts as a `controller' that can prime System 1 to appropriately bias its fast thinking. In the context of reasoning with deep learning models, System 2 has been interpreted as operating with sparse concepts that can be described in language \citep{bengio1,bengio2}. Through repeated usage, the functions of System 2 become compressed into System 1 intuitions, in the same manner that iterative `reasoning' functions of which smaller LLMs are not capable become zero-shot generation capacities for large LLMs. 
As is the case with humans, there is always the next frontier of problems where a trained model with remarkable `intuition' needs to be slowed down. The main claim of this paper is that more is possible with LLMs of existing scale when they are used in concert with a wise controller that allows for probabilistic inference.

\section{ThinkSum}
\label{sec:thinksum}

\subsection{How to \Think{}} %

Here we list examples of the ``fast thinking" that precedes the summarization stage.

\paragraph{Elementary string manipulations.}
Standard ways to turn a question into a prompt that can be given to a LLM for generation or scoring involve choices (e.g., of the prompt format) that can be seen as being made by a controlling agent. The default approach to multiple-choice questions is to write them as Cloze tasks. However, there are nontrivial operations used in inference procedures that sometimes work better, such as:
\begin{enumerate}[left=0pt,label=$\bullet$,nosep]
\item \thinktechnique{Order inversion}: Exchanging the order of the question and answers, as in \citet{min-etal-2022-noisy}.
\item \thinktechnique{Premise erasure}: Deleting a part of the question. Removing a premise with which the answer is expected to have high mutual information is a step in inference procedures that aim to correct for bias towards answers with high unconditional likelihood \citep{zhao2021calibrate,holtzman-etal-2021-surface,whitammer-etal-2022-coherence}.
\end{enumerate}

\paragraph{\thinktechnique{Substitution and normalization}.}

An example is shown in Fig.~\ref{fig:thinksum_teaser}. Elements from a set may be substituted in place of `slot' words in a prompt, such as `cat' substituted for `binne' in the prompt ``\texttt{A binne bam is a place for}''. This operation can be combined with syntax-normalization steps that are reliably achieved by standard NLP tools, such as ensuring subject-verb agreement. 

\paragraph{Example and list generation.}

A LLM can be prompted to generate or score lists of words or phrases. We suggest and experiment with three instances of this:
\begin{enumerate}[left=0pt,label=$\bullet$,nosep]
\item \thinktechnique{Example generation}: In Fig.~\ref{fig:thinksum_teaser}, the LLM is prompted to turn a definition or characterizing property, such as `simple dwelling', into a list of examples. This can be achieved with a prompt such as ``\texttt{A bam is a simple dwelling. Examples: 1.}''. The generated completion can be parsed into a set to be used later in the inference procedure. 
\item \thinktechnique{List extension}: A similar approach can also be used to hallucinate additional possible answers to questions, as we will show in some of the experiments.
\item \thinktechnique{List of words}: Similar prompts provide an even simpler \Think{} method that we use for scoring -- but not generation -- in several tasks. Just prompting a LLM with ``\texttt{List of words: $A$, $B$}'', where $A$ and $B$ are words or phrases, and computing the likelihood of $B$ conditioned on ``\texttt{List of words: $A$,}'' is a good measure of semantic relatedness of $A$ and $B$.
\end{enumerate}

\paragraph{\thinktechnique{Fact generation}.}

This way of \Think{ing} associates an input word with a set of phrases in a similar manner to generating examples from a definition. It can be achieved with prompts such as ``\texttt{List facts about cats. 1.}'' The generated facts are good targets for substitutions of other concepts (`dogs', `galaxies') in place of the concept (`cats') about which facts are generated. A variation on this asks the LLM to generate differences between two concepts, as shown in Fig.~\ref{fig:odd_combined} (right).

\paragraph{\thinktechnique{Translation}.}

The LLM can be prompted to convert between different forms of representing the same concept as a sequence of tokens. We use two basic examples of this in experiments:
\begin{enumerate}[left=0pt,label=$\bullet$,nosep]
\item Translation between languages by prompting the LLM in formats such as ``\texttt{French: J'adore les chats noirs. English:}''. A very similar approach can be used to convert non-alphabetic symbols, such as emoji, into words with similar meanings.
\item Converting text to formal (symbolic) structures, like turning a word problem into a collection of mathematical equations.
\end{enumerate}

\subsection{How to \Sum{}} %

\paragraph{Elementary inference.}

As above, we begin by listing existing standard ways of turning LLM outputs into answers, which we see as trivial cases of aggregation (\Sum{}).
\begin{enumerate}[left=0pt,label=$\bullet$,nosep]
\item \sumtechnique{Majority/minority vote (argmin/argmax)}: a component of most answer selection procedures.
\item \sumtechnique{Ratio of likelihoods}: Likelihoods from different variants of the same prompt can be combined by considering their ratio or more general log-linear or other mixture. For example, this can be done to correct the likelihood of an answer conditioned on a question by its unconditional likelihood, in combination with the \thinktechnique{Premise erasure} operation described above.
\end{enumerate}

\paragraph{\sumtechnique{Mixture (average) aggregation}.}
A collection of prompts can be treated as the components of a mixture model over completions. An example is shown in Fig.~\ref{fig:thinksum_teaser}, where substitutions of a set of words yield 25 different prompts. Likelihoods of the completion over these 25 prompts are averaged.

\paragraph{\sumtechnique{Product aggregation}.}
We use products of likelihoods in two different ways:
\begin{enumerate}[left=0pt,label=$\bullet$,nosep]
\item In a similar way as mixtures, but when the more natural probabilistic model has \emph{all} elements of a set (of prompts) generating the answer, such as when a description or definition must be satisfied by all concepts in a set.
\item In a task where we are to determine whether a statement $S$ or its negation $S'$  is true, we can compute the likelihood of both $S$ and $S'$ being true (as posterior over the tokens `True' and `False' in an appropriate prompt), then compare $p(\texttt{True}|S)p(\texttt{False}|S')$ ($S$ is true and $S'$ is false) with $p(\texttt{False}|S)p(\texttt{True}|S')$ ($S$ is false and $S'$ is true).
\end{enumerate}

\section{Experiments}

In this section, we perform case studies on three tasks from the BIG-bench suite to demonstrate the possibilities of the inference approaches discussed in \S\ref{sec:thinksum}.  We also experiment with ten other tasks from BIG-bench; the best results are summarized in Table~\ref{tab:results_main} and the methods, grouped by the style of \Think{ing} and \Sum{ming}, are described in Appendix (\S\ref{sec:add_exp}).

All details of the tasks can be found in the Appendix (\S\ref{sec:app-tasks}). Comparisons to direct prompting and algorithms that append retrieved or generated tokens to the prompt are given in \S\ref{sec:thinksum-vs-cot}.

\begin{table*}[t]
\centering
\resizebox{\linewidth}{!}{
\begin{tabular}{@{}lcccccccc}
\toprule
&& \multicolumn{4}{c}{GPT-3 (davinci) $n$-shot} & \multicolumn{3}{c}{\ThinkSum} \\
\cmidrule(lr){3-6}\cmidrule(lr){7-9}
\textbf{Task} & Avg. H & $n=0$ & 1 & 2 & 3 & GPT-3 & InstructGPT & GPT-2 XL \\
\midrule
\taskname{Invented words} (\S\ref{sec:exp_binnebam}) & N/A & 0.29 & 0.14 & 0.14 & 0.21 & 0.64 & \textbf{0.71} & 0.29 \\
\taskname{Odd one out} (\S\ref{sec:exp_ooo}) & 0.80 & 0.27 & 0.20 & 0.23 & 0.23 & 0.80 & \textbf{0.84} & 0.71 \\
\taskname{Five objects} (\S\ref{sec:exp_logical}) & N/A & 0.23 & 0.29 & 0.28 & 0.32 & -- & \textbf{0.77} & -- \\ \midrule
\taskname{Sports understanding} (\S\ref{sec:exp_uncertainty}) & 0.71 & 0.50 & 0.50 & 0.50 & 0.50 & 0.71 & \textbf{0.74} & 0.54 \\
\taskname{Known unknowns} (\S\ref{sec:exp_uncertainty}) & \textbf{0.80} & 0.61 & 0.52 & 0.48 & 0.50 & 0.54 & 0.76 & -- \\
\taskname{Misconceptions Russian} (\S\ref{sec:exp_translation}) & 0.65 & 0.33 & 0.33 & 0.41 & 0.35 & \textbf{0.70} & 0.61  & --\\
\taskname{Emoji movie} (\S\ref{sec:exp_translation}) & \textbf{0.93} & 0.12 & 0.18 & 0.12 & 0.19 & 0.80 & 0.75 & -- \\
\taskname{ParsiNLU reading comprehension} (\S\ref{sec:exp_translation}) & 0.02 & 0.00 & 0.00 & 0.00 & 0.00 & -- & 0.02 & -- \\
\taskname{Phrase relatedness} (\S\ref{sec:exp_relatedness}) & 0.74 & 0.37 & 0.42 & 0.52 & 0.59 & 0.85 & \textbf{0.87} & 0.79 \\
\taskname{Codenames} (\S\ref{sec:exp_relatedness}) & 0.18 & 0.01 & 0.11 & 0.16 & 0.19 & 0.37 & \textbf{0.41} & 0.36 \\
\taskname{Novel concepts} (\S\ref{sec:exp_substitution}) & 0.67 & 0.47 & 0.47 & 0.56 & 0.56 & 0.72 & \textbf{0.75} & 0.50 \\
\taskname{Code line description} (\S\ref{sec:exp_substitution}) & 0.60 & 0.32 & 0.32 & 0.28 & 0.32 & 0.83 & \textbf{0.90} & 0.77\\ %
\taskname{Language identification} (\S\ref{sec:exp_other}) & 0.16 & 0.16 & 0.12 & 0.13 & 0.11 & \textbf{0.57} & -- & 0.30  \\
\bottomrule
\end{tabular}
}
\caption{Standard metric (BLEU for \taskname{Codenames}, accuracy for other tasks) for GPT-3 175B (davinci) and \ThinkSum{} with 175B (davinci), InstructGPT and GPT-2 XL on BIG-bench tasks. A `--' indicates that the model and task combination was not evaluated because the model does not reliably execute the appropriate \Think{} prompt. We did not evaluate InstructGPT on \textsc{Language identification} due to the large dataset size and API quota.}
\label{tab:results_main}
\end{table*}

\subsection{Conceptual combinations: Invented words}
\label{sec:exp_binnebam}

In \taskname{Invented words}, two nonce words $x_1,x_2$ are defined and the correct statement must be chosen out of a set of statements $S=\{s_j\}$ that begin with (possibly inflected forms of) ``$x_1$ $x_2$'' (Fig.~\ref{fig:thinksum_teaser}).

We use an \thinktechnique{Example generation} prompt to obtain a set of example words fitting the definitions of $x_1$ and $x_2$. We thus obtain sets $S_1$ and $S_2$ of words that can be substituted for $x_1$ and $x_2$, respectively.

We treat each statement $s_j$ as a template into which words $w_1\in S_1$ and $w_2\in S_2$ can be substituted by replacing $x_i$ with $w_i$ and normalizing the syntax to ensure subject-verb agreement. Denoting by $s_j\langle w_1,w_2\rangle$ such a substitution, we form a vector of probabilities $p_j$ by scoring the \thinktechnique
{Substitution} of each possible pair of words into each statement and performing \sumtechnique{Mixture aggregation} and considering the \sumtechnique{Ratio of likelihoods} with the template without substitution:
\begin{equation*}
    p_j=\frac{\frac{1}{|S_1||S_2|}\sum_{w_1\in S_1,w_2\in S_2}p_{\LM}(s_j\langle w_1,w_2\rangle)}{p_{\LM}(s_j)}.
\end{equation*}
The statement $s_j$ with highest likelihood under this normalized mixture, $\argmax_j p_j$, is selected.%

\subsection{Odd one out} 
\label{sec:exp_ooo}

\begin{figure*}[t]
\includegraphics[width=0.4\textwidth,height=13em]{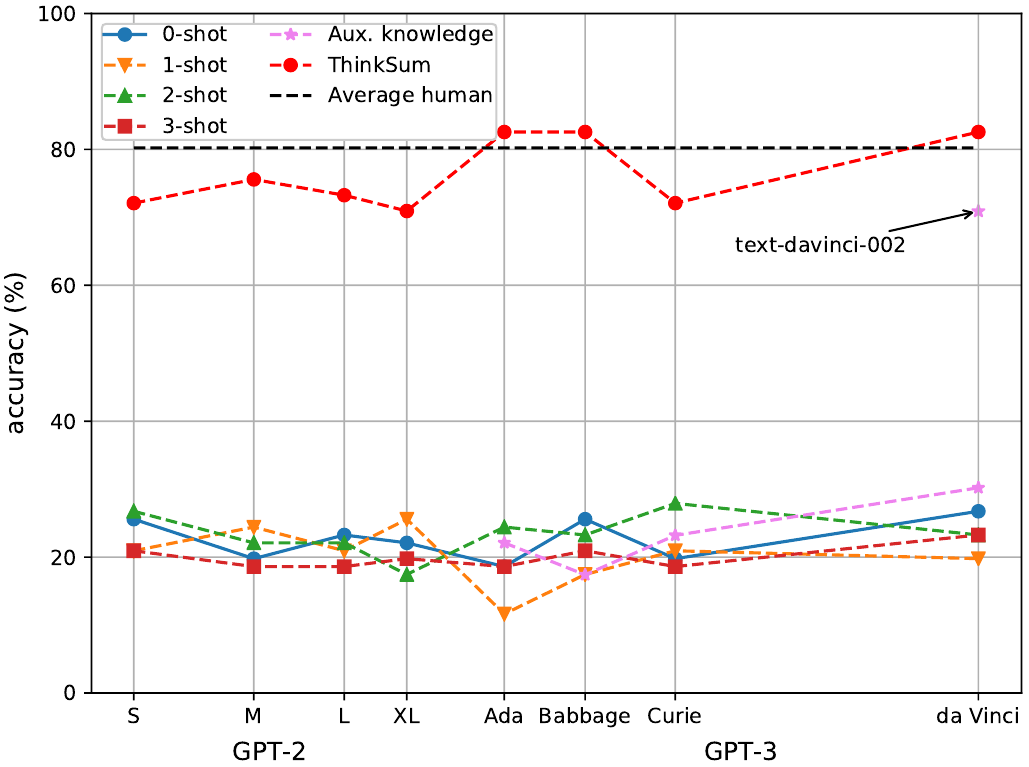}~
\includegraphics[width=0.26\textwidth,height=12.5em,trim=0 0 0 5,clip]{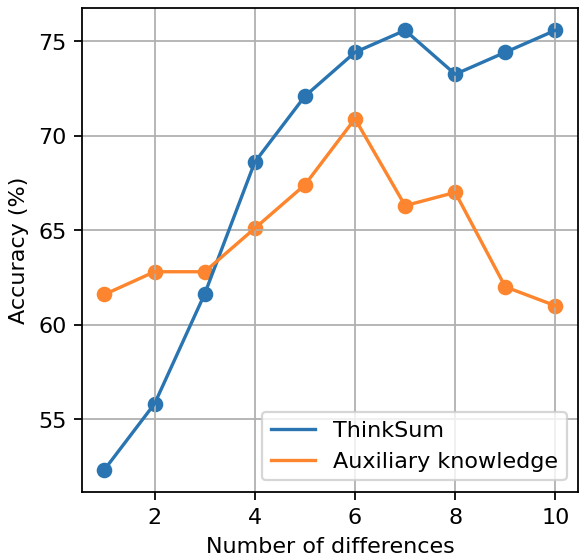}~\hspace{1em}~
\raisebox{1em}{\includegraphics[width=0.28\textwidth,height=12em,trim=0 0 0 8,clip]{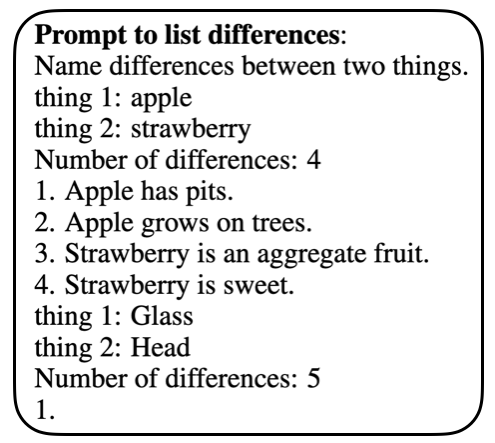}}
\caption{\taskname{Odd one out}. \textbf{Left:} Performance of GPT-3 ($n$-shot, $n = 0,1,2,3$), auxiliary knowledge, and \ThinkSum{} with various model sizes. \textbf{Middle:} Auxiliary knowledge vs. \ThinkSum{} with varying number of differences. \textbf{Right:} Prompt used to generate knowledge statements.}
\label{fig:odd_combined}
\end{figure*}

We examine possible \Think{} and \Sum{} approaches in depth on the \taskname{Odd one out} task, in which the word in a set $W=\{w_i\}$ that is \emph{least} semantically related to the others must be chosen (e.g., \textit{Pick the odd word out: glass, head, arm, leg, hand, foot}).

\paragraph{List of words.} We form a semantic relatedness matrix $P_{ij}$ by querying the LLM with a \thinktechnique{List of words} \Think{} prompt for each pair of indices $i,j$:
\[P_{ij}=p_{\LM}(w_j\mid\text{``\texttt{List of words: $w_i$, }''}).\]
This matrix is aggregated by averaging over $j$ (in log domain) and selecting the $i$ with lowest average, i.e., least likelihood of being generated by a product mixture of all words in the set: $i=\argmin_i\prod_j P_{ij}$. This is a case of \sumtechnique{Product aggregation}.

Because this approach is the most successful with all model sizes we experimented with, its performance is reported in Table~\ref{tab:results_main}. Remarkably, near-average-human accuracy is maintained for all model sizes from GPT-2 Small to the largest GPT-3 model (Fig.~\ref{fig:odd_combined} (left)).

\paragraph{Fact generation.} As an alternative approach, we use a \thinktechnique{Fact generation} prompt. An effective way to mine facts for semantic relatedness tasks is to consider two items in the same context in order to get relevant facts regarding how items are related to each other (prompt in Fig.~\ref{fig:odd_combined} (right)). The demonstration used in the prompt ensures that the LLM generates statements in an expected format, which can be parsed and used for probability computation later. Using this prompt, we obtain a collection of statements $S=\{s_i\}$ about items $w_j$. We treat each generated $s_i$ as a template into which different words $w$ can be substituted and denote by $s_i\langle w\rangle$ the \thinktechnique{Substitution} of word $w$ into template $s_i$. We then form a $|S|\times|W|$ matrix $P_{ij}$, defined by 
$P_{ij}=p_{\LM}(s_i\langle w_j\rangle)$. Then, we can perform \sumtechnique{Minority voting}: we take argmin over $j$ and pick as the answer the most frequently occurring value, i.e., the item that is most often the least likely to fit a generated statement.

\paragraph{Comparison with auxiliary knowledge approaches.} 

We compare our method with a knowledge-based prompting method, herein referred to as auxiliary knowledge. In auxiliary knowledge, we prepend generated facts in the prompt before the question. Details of the prompt for auxiliary knowledge are provided in \S\ref{sec:know}. 
In Figure \ref{fig:odd_combined} (middle), we show that the accuracy of \thinktechnique{Fact generation}-based \ThinkSum{} rises as the number of generated facts is increased, while the auxiliary knowledge technique peaks and then degrades as the prompt lengthens.

Fig.~\ref{fig:odd_combined} (left) shows how performance varies with the size of the LLM used for GPT-3, auxiliary knowledge and \ThinkSum{} on \taskname{Odd one out}. Even with GPT-2 Small, \ThinkSum{} dramatically improves over much larger largest zero- or few-shot models with or without auxiliary knowledge. A finetuned iteration of the largest GPT-3 model, text-davinci-002, is the only model variant that, with the help of auxiliary knowledge, achieves competitive performance with \ThinkSum{}. This result provides experimental evidence for our claim that while new models may create qualitative jumps, \ThinkSum{} can push the performance limits of smaller models.

\paragraph{Latent variable models.}  As we have shown, the detection of the odd item can be performed with simple inference operations on items, facts, and their joint likelihoods. However, it is also possible to assume a latent structure in the items and facts, consisting of two or more clusters such that the facts and items belonging to a cluster can be freely interchanged. We describe a problem-specific latent variable model that enables selecting the facts that characterize the majority class, thus explaining why the minority item is ruled as the odd one out and helping interpret the decisions of the system.  

We model items $i \in I$ and facts $f \in F$ as being generated from a latent class $c \in \{0,1\}$. The  distribution is modeled as:
\begin{align*}
    P(i,f)=\sum_c P(c)P(i|c)P(f|c)
\end{align*}
where $P(i,f)$ is a matrix of likelihoods from the LLM and the semantic components, groupings $P(i | c)$ and $P(f | c)$, are derived from the matrix using a standard iterative expectation-maximization \cite[EM;][]{EM} inference procedure (see \S\ref{sec:em}). 
Then, the score for an item $i$ belonging to a cluster and all other items $m \in S, \{m \neq i\}$ belonging to another cluster can be found as $S_i = \sum_{c, c' \neq c}P(i|c)P(c)\prod_{m \neq i}P(m|c')P(c')$.

We show the effectiveness of the latent variable models in Table~\ref{tab:sum_options}, where we analyze different methods for solving \taskname{Odd one out} using the InstructGPT variants text-davinci-001 and text-davinci-002. For the `latent variable model' and `minority voting' methods, we use number of differences $N_d = 5$. The latent variable model is trained for $200$ EM iterations. All probabilistic reasoning methods perform well, outperforming previous baselines reported in Table \ref{tab:results_main}. Inference using EM, as well as the other approaches, can be seen as a \Sum{} (inference) operation and can be applicable in other tasks of similar structure.

\begin{table}[t]
\centering
\begin{tabular}{@{}lcccc}
\toprule
            Model &  LoW & LVM & MV  \\
                \midrule
text-davinci-002  & 0.84 & 0.67 & 0.70 \\ 
text-davinci-001  & 0.74 & 0.77 & 0.70 \\\bottomrule
\end{tabular}
\caption{Different alternatives of probabilistic reasoning with \ThinkSum{} for solving \taskname{Odd one out}: list of words, latent variable model, minority voting.}
\label{tab:sum_options}
\end{table}

\subsection{Logical deduction}
\label{sec:exp_logical}

\begin{figure*}[t]
  \centering
  \begin{center}
      \includegraphics[width = 0.9\textwidth,trim=0 10 0 0]{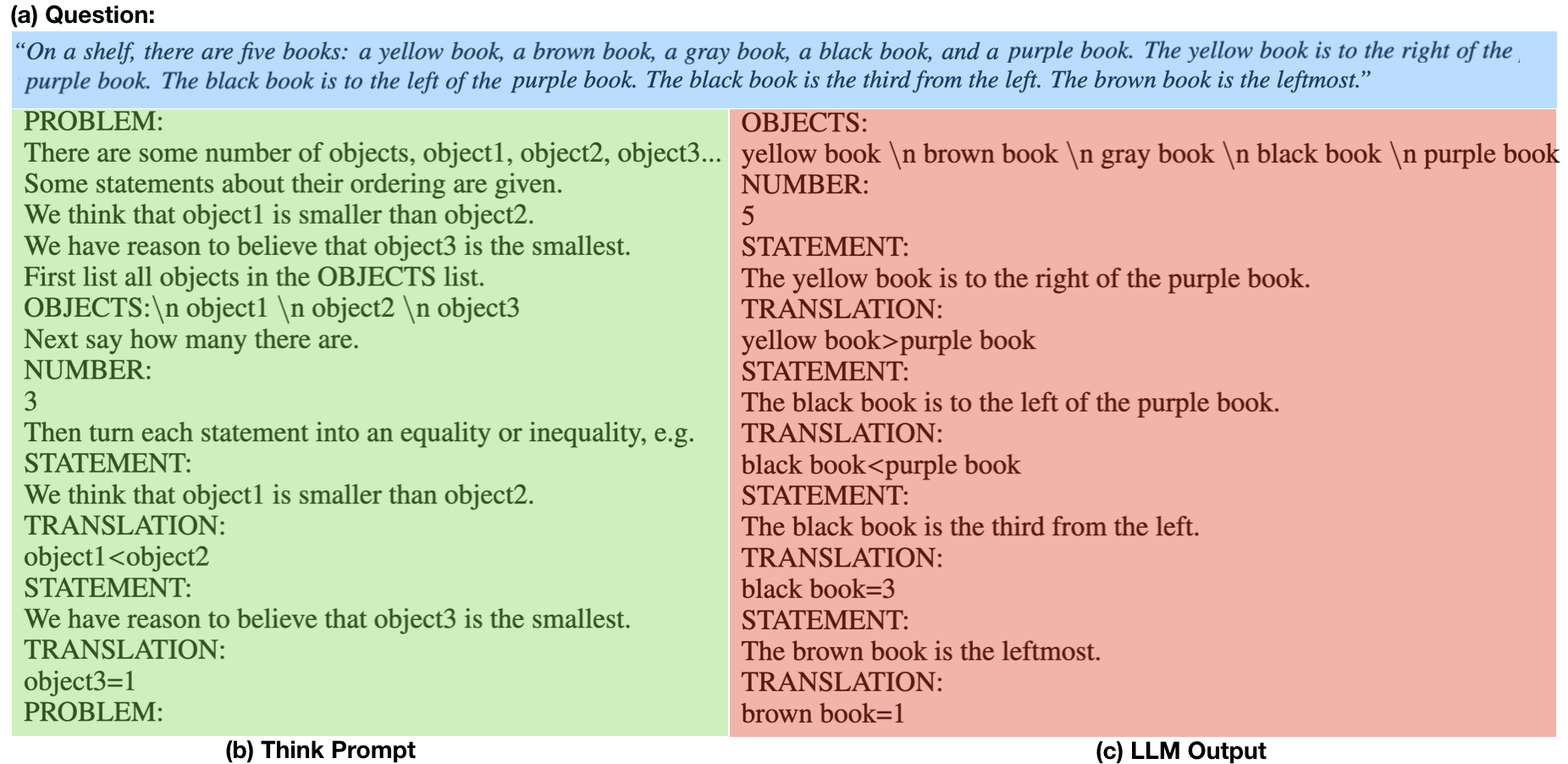}
  \end{center}
  \caption{Details for \taskname{Logical deduction}. (a) Example question from the task, (b) demonstration for the \Think{} prompt, (c) example LLM output. The demonstration induces the LLM to generalize from generic objects ordered by size to books ordered by position.}
 \label{fig:logical}
\end{figure*}

In the \taskname{Logical deduction} task, different types of items and clues regarding their order are provided (Fig.~\ref{fig:logical}(a)). The goal is to select the correct statement from a set of statements about their placements. The ordering problems involve different types of objects (cars, birds, etc.) and orderings (by size, price, contest ranking, etc.). The task creators emphasize that this task requires parsing information about multiple objects and their relationships, understanding rules regarding ordered objects in various scenarios, and iteratively applying these rules. The LLM calls in the \Think{} stage of \ThinkSum{} can perform mappings required to parse information and understand rules, and the \Sum{} stage can integrate mappings of objects to the placements under these rules. Here, we use a \thinktechnique{Translation} prompt to map the given problem into a set of mathematical (in)equalities (Fig.~\ref{fig:logical}(c)).

The \thinktechnique{Translation} prompt in Fig. \ref{fig:logical}(b), containing generic ordering statements and object names that are not used in the task as an in-context demonstration, is sufficient to perform the translation from natural language to equations. By prepending this demonstration prompt to a problem statement, we induce the LLM to map the objects in the problem to the set of strings corresponding to numbers from $1$ to $N$, where $N$ is the number of objects, and to produce a set of inequalities (Fig.~\ref{fig:logical}(c)).

Once a translation of the problem into a set of inequalities is obtained,  the \Sum{} stage considers all possible mappings of items to indices to determine the mapping compatible with the discovered set of (in)equalities. This can be done by an external algorithm or by the LLM itself, as an LLM may be capable of understanding that, for example, ``\texttt{2>3}'' is a less likely string than ``\texttt{2>1}'' (see \S\ref{sec:llm_ineq}). 

Finally, the probability of each of the candidate statements, like ``\texttt{yellow book=2}'', can thus be obtained by: 
\begin{align}
    p(``\texttt{yellow book=2}''&\mid T)\nonumber\\
    \propto\sum_{\textbf{b}\in \{1,\dots,N\}^N} p_{\LM}(&\{T_t\langle\textbf{b}\rangle:T_t\in T\}\label{eq:ineq_sum}\\
    &\cup\{``\texttt{yellow book=2}''\langle\textbf{b}\rangle\})\nonumber
\end{align}
where $\textbf{b}$ denotes the vector of positions for the $N$ items (e.g., $(5,2,3,4,1)$), $T=\{T_t\}_{t=1}^N$ is the set of inequalities obtained from the \thinktechnique{Translation} prompt as a set of strings (e.g., ``\texttt{black book<purple book}''), and $s\langle\textbf{b}\rangle$ denotes the substitution of the corresponding entry in $\textbf{b}$ in place of the object name in the string $s$ (e.g., ``\texttt{4<5}''). The term inside the sum is a case of \sumtechnique{Product aggregation}: the LLM likelihoods of all strings in the set are multiplied.

In summary, our solution to this task involves composition of two \Think{} operations -- a \thinktechnique{Translation} into a set of equations and then \thinktechnique{Substitution} of numbers in place of item names -- and two \Sum{} operations -- a \sumtechnique{Product aggregation} followed by a \sumtechnique{Mixture aggregation}. (Other options are discussed below.)

\paragraph{Results and discussion.}
For the 500 \taskname{Logical deduction} problems with $N=5$ objects, \ThinkSum{} yields an accuracy of $77\%$ (see Table \ref{tab:results_main}), besting the average human performance. When the necessary summations become large, it becomes very unlikely that pure prompt engineering can be competitive, as even humans need paper and pencil to create and attend to many alternative solutions, and would likely translate the premises into a simpler notation using a single letter (representing a variable to which a numeric value can be assigned) to represent each object, rather than directly attending to the words in the problem statement. 

We also test an auxiliary knowledge method akin to chain-of-thought reasoning, where the information obtained with the prompt in Fig.~\ref{fig:logical} is appended to the LLM input. In particular, the problem, together with its translation into inequalities, is used as a prompt to each of the answer options, and then the option with the highest likelihood is chosen for the answer. This approach does improve over straightforward zero-shot GPT-3 scoring, but only raises the accuracy to $50\%$ (see \S\ref{sec:thinksum-vs-cot} and Table \ref{tab:thinksum_vs_auxiliary}).

\paragraph{Optimizations, failure modes, and extensions.} We have seen that InstructGPT is able both to translate logical deduction problems into (in)equalities (Fig.~\ref{fig:logical}) and to evaluate each of them after replacement of items with position numbers (\S\ref{sec:llm_ineq}). We conclude that the \Sum{} stage is there simply to \emph{search} over all possible mappings, the way a human might. But, just as a human might use shortcuts in the search, the \Sum{} stage of \ThinkSum{} could be implemented in more or less efficient ways. For example, instead of summing over all possible assignments of the five items, we can avoid the ones that are not permutations of $\{1,2,3,4,5\}$. Furthermore, instead of using $p_{\text{LLM}}$ from Fig.~\ref{fig:inequality_eval} in (\ref{eq:ineq_sum}), we can simply evaluate each inequality externally, giving a high constant probability for each inequality $T_t\langle\textbf{b}\rangle$ that is true and a low probability when it is false, or the summing can be aborted whenever an incorrect statement is detected in a particular assignment \textbf{b} of positions to items.

The prompt in Fig.~\ref{fig:logical}(b) instructs the LLM to assign positive integers depending on the language used (e.g., the smallest object gets 1), but a common behaviour of the LLM is to generalize to assigning negative numbers, such as using $-2$ to represent `second from the end' (or second-largest, etc.). To remain robust to such a behavior of the \Think{} stage, we can convert negative position numbers $r$ into $N+r+1$ before evaluating statements. However, a persistent failure mode of this kind of \ThinkSum{} is that the LLM may translate inequality statements inconsistently with equality statements (e.g., by coding the leftmost item as 1 and being consistent with this choice for other equality constraints, but translating inequality constraints consistently with the reverse order, with `left of' meaning $>$). Such failures can be addressed by careful engineering in the \Sum{} stage, such as by summing out a binary latent variable indicating whether inequalities should be reversed. This increases the number of model evaluations, but also allows for robust auto-correction by the \Sum{} stage of inconsistencies in the \Think{} stage.

\subsection{Comparisons with chain-of-thought and auxiliary knowledge approaches}
\label{sec:thinksum-vs-cot}

\paragraph{\ThinkSum{} vs.\ auxiliary knowledge.} Table \ref{tab:thinksum_vs_auxiliary} shows the comparison of \ThinkSum{} with algorithms that append auxiliary knowledge as an oracle `reasoning chain'. For \taskname{Phrase relatedness}, auxiliary knowledge was generated using the ``list differences'' prompt shown in Fig.~\ref{fig:odd_combined} (right). For both auxiliary knowledge and \ThinkSum{}, 6 generated differences were used, as that was the best for auxiliary knowledge (see Fig.~\ref{fig:odd_combined} (middle)). \ThinkSum{} \taskname{Odd one out} and \taskname{Phrase relatedness} are solved with the ``list of words'' prompt. For \taskname{Logical deduction}, the \Think{} prompt shown in Fig.~\ref{fig:logical} was included before the question in the prompt. In all cases, \ThinkSum{} outperforms auxiliary knowledge. 

\begin{table}[t]
\centering
\resizebox{1\linewidth}{!}{
\begin{tabular}{@{}lccc}
\toprule
                & \taskname{Odd one out}   & \taskname{Phrase relatedness} & \taskname{Logical deduction ($N=5$)} \\ \midrule
\ThinkSum{}  & 0.84 & 0.87               & 0.77                     \\
Aux.\ knowledge &       0.71     &    0.75                &                  0.50        \\ \bottomrule
\end{tabular}
}
\caption{\ThinkSum{} vs.\ auxiliary knowledge with text-davinci-002.}
\label{tab:thinksum_vs_auxiliary}
\end{table}

\paragraph{\ThinkSum{} vs.\ chain of thought.}

Following \citet{wei2022chain}, we use ``chain-of-thought (CoT) methods" to mean LLM scoring approaches that use insertion of generated tokens between the prompt and the target answer. The model is taught, using few-shot demonstrations, how to generate these intermediate tokens. Above we have compared \ThinkSum{} with approaches that add \emph{extracted} (from an auxiliary LM call), not \emph{generated} (within the LM's linear workspace) token sequences after the prompt, for the \taskname{Odd one out}, \taskname{Phrase relatedness}, and \taskname{Logical deduction} tasks (see Table \ref{tab:thinksum_vs_auxiliary}).

With suitable examples, it may be possible for a CoT approach to replace the \Think{} phase, by learning from demonstrations to generate the appropriate knowledge, and parts of the \Sum{} phase, although inference over parallel evaluations of the LLM is no longer possible. Our auxiliary knowledge baselines make precisely that generous assumption and focus the comparisons on the need for parallel calls and reasoning over possibilities using probabilistic inference (instead of leaving it to the LLM to make the right conclusions from the list of extracted alternatives).

Although we expect that appending facts in a standard format to the prompt would help the model more than teaching the model to generate these facts, we experimented with CoT approaches on several tasks. Table \ref{tab:cot_prompt_examples} shows example demonstrations and prompt formats used for each task, and Table \ref{tab:cot_results} shows the results using two variants of the largest GPT-3 model.

\begin{table}[t]
\centering
\resizebox{1\linewidth}{!}{
\begin{tabular}{lccccc}
\toprule
& \multicolumn{3}{c}{GPT-3 (davinci)} &\multicolumn{2}{c}{GPT-3 (davinci-002)}\\\cmidrule(lr){2-4}\cmidrule(lr){5-6}Task & Direct & CoT & \ThinkSum & CoT & \ThinkSum\\\midrule
\taskname{Odd one out} & 0.27 & 0.33 & 0.80 & 0.64 & 0.84 \\
\taskname{Phrase relatedness} & 0.59 & 0.55 & 0.85 & 0.79 & 0.87 \\
\taskname{Logical deduction} & 0.32 & 0.25 & -- & 0.39 & 0.77 \\
\taskname{Known unknowns} & 0.61 & 0.70 & 0.54 & 0.74 & 0.76 \\
\taskname{Invented words} & 0.29 & 0.50 & 0.64 & 0.64 & 0.71
\\\bottomrule
\end{tabular}
}
\caption{Comparison of \ThinkSum{} with chain-of-thought prompting approaches.}
\label{tab:cot_results}
\end{table}

As expected, \ThinkSum{} outperforms CoT prompting on all tasks with all variants except \taskname{Known unknowns} with the davinci variant, where direct prompting already performs well. (We did not evaluate \ThinkSum{} with davinci on \taskname{Logical deduction} because prompts like the one in Figure~\ref{fig:logical} did not reliably produce outputs in the correct format; notice that CoT is barely better than random guessing (20\%).) 

When interpreting these results, it is important to note that only one prompt format was evaluated for both CoT and \ThinkSum{}, and the format of prompts and demonstrations can have a strong and often unpredictable effect on the LLM. We observed that CoT approaches are highly sensitive to minor changes in the prompt format or the construction of in-context examples, consistent with the known biases of in-context learning \citep{lu-etal-2022-fantastically,zhao2021calibrate}. On the other hand, using structured, shorter components is more reliable, as demonstrated by the efficacy of the \Think{} prompts used in \ThinkSum{}.

\section{Related work}

\paragraph{Improvements to LLM inference.}

After the discovery of the in-context learning abilities of LLMs, there has been an explosion of interest in improving inference with LLMs in the zero-shot and few-shot setting \cite{gpt3,chowdhery2022palm, rae2021scaling}. One approach to improving the reasoning abilities of LLMs involves appending, or learning to generate, auxiliary knowledge within the prompt \citep{shwartz-etal-2020-unsupervised,zelikman2022star,nye2021show}. Recently, more general auxiliary knowledge or chain-of-thought prompting methods have been proposed \citep{wei2022chain, wang2022self, zhou2022least, creswell2022selection, wang2022rationale,liu-etal-2022-multi}, including those that allow a control flow external to the main LLM \citep{decomp}. Later, \citet{kojima2022large} showed zero-shot chain-of-thought prompting can improve performance on a variety of reasoning tasks. This method does not require any hand-crafted few-shot examples, which is a shared property with \ThinkSum{}. 
\citep{Nye2021sequence} observed that a dual-system approach where an associative ``System 1'' and a logical ``System 2'' can increase coherence of LLMs in tasks such as robust story generation and grounded instruction following. The two-step paradigm in \ThinkSum{} is similar, where ``System 1'' is the (querying of the LLM for) fast thinking, and ``System 2'' is the probabilistic inference step.

\paragraph{Brittleness of chain-of-thought prompting.} Despite the recent success of chain-of-thought approaches, recent studies have raised concerns regarding the limitations of chain-of-thought approaches. \citet{webson-pavlick-2022-prompt} observed that instructive prompts perform similarly with misleading or intentionally irrelevant prompts. Additionally, \citet{ye2022unreliability} showed improvements due to few-shot chain-of-thought are not observed in question answering, or natural language inference. More critically, few-shot prompts are highly sensitive to the order in which the samples are provided, the prompt format, and the selection of in-context examples, \citep{lu-etal-2022-fantastically,zhao2021calibrate}. Thus, it is crucial to design techniques that are robust to such changes in the prompt.
\paragraph{Inference as reasoning.}
Iterative inference over LLM outputs has been proposed for tackling true/false question answering and commonsense question answering \citep{jung2022maieutic,liu-etal-2022-generated}. \citet{xie2021explanation} presents a Bayesian inference perspective on in-context learning, and \citet{dohan2022language} formalizes and unifies existing prompting techniques in a probabilistic framework. Our work generalizes such approaches to perform arbitrary probabilistic inference outside of the LLM.

\section{Conclusion}
In this paper we presented \ThinkSum{}, a two-step probabilistic inference paradigm that reasons over sets in a structured manner. The fast thinking stage of \ThinkSum{} allows elementary string manipulations as well as natural language prompting, which may enable numerous approaches to solve a natural language task. Even with far smaller model variants, \ThinkSum{} achieves state-of-the-art results on ten difficult tasks in BIG-bench using GPT-family models. The two-step paradigm allows operating over sets instead of manipulating the prompt itself, preventing sensitivity to prompt format during the probabilistic inference in \ThinkSum{}, which is performed outside of calls to the LLM. As a result, \ThinkSum{} is more robust to prompt design, yields more interpretable predictions, and can be combined with many probabilistic inference approaches to tackle a diverse set of tasks.

 \section*{Acknowledgments}

 The authors thank Alexandros Graikos, Sudha Rao, and Alessandro Sordoni for valuable discussions.

\section*{Limitations}

Our proposed \ThinkSum{} has demonstrated strong performance on thirteen challenging BIG-bench tasks. However, it is important to acknowledge certain limitations of the system.

Firstly, as the number of objects or facts that are reasoned over increases, the computation cost will also rise. However, increasing the number of objects will also make the task harder, and direct prompting may cease to work at all (as we indeed observe in BIG-bench results, such as \taskname{Logical deduction} with more than five objects), while \ThinkSum{} offers a generalizable methodology, as the atomic \Think{} operations do not increase in complexity as the number of objects grows.%

Secondly, when solving a new task, it is necessary to expend human effort to select specific operations in each step, as outlined in \S\ref{sec:thinksum}. This limitation is shared with prompt engineering of all kinds, including direct or chain-of-thought prompting: finding a prompt for a new task requires an often-cumbersome prompt engineering procedure. We have described \ThinkSum{} as a general two-stage paradigm, with an external inference step. This generality aims to facilitate the adaptation of \ThinkSum{} to new tasks, with minimal modifications to the \Think{} and \Sum{} steps. Work on automating the prompt engineering procedure \citep{ape} is a promising path towards overcoming this limitation. An alternative to prompt engineering that does not require such human effort is tuning (i.e., differentiable end-to-end learning) of prompts or model parameters; however, this remains impractical for GPT-3-scale models, and attempts to tune models directly on symbolic reasoning chains have met with limited success \citep{kassner-etal-2020-pretrained}. 

Last but not least, \ThinkSum{} has mainly been evaluated with GPT-3 (davinci) and InstructGPT (text-davinci-002) models. To further improve performance, it may be beneficial to apply \ThinkSum{} to more recent instruction-tuned models such as Flan-PaLM~\cite{chowdhery2022palm, chung2022scaling}, text-davinci-003, ChatGPT, and GPT-4, which seem more capable of robustly performing \Think{} steps.%

\section*{Ethics and impact statement}

We foresee no direct or immediate societal impacts arising from this work. However, we would like to emphasize that relying solely on LLMs' associative reactions to prompts can lead to undesired bias in the behaviour of systems. Control of LLMs' reasoning in the way we have proposed can potentially mitigate such bias, due both to the decomposition of the argumentation process into interpretable fact-retrieval steps and to the averaging effect of smoothing out spurious triggers when aggregating many hypotheses and reasoning chains.

\bibliography{anthology,custom}

@inproceedings{whitammer-etal-2022-coherence,
    title = "Coherence boosting: When your pretrained language model is not paying enough attention",
    author = "Whitammer, Esmeralda S.  and
      Wang, Zhen  and
      Jojic, Nebojsa",
    editor = "Muresan, Smaranda  and
      Nakov, Preslav  and
      Villavicencio, Aline",
    booktitle = "Proceedings of the 60th Annual Meeting of the {A}ssociation for {C}omputational {L}inguistics (Volume 1: Long Papers)",
    month = may,
    year = "2022",
    address = "Dublin, Ireland",
    publisher = "Association for Computational Linguistics",
    url = "https://aclanthology.org/2022.acl-long.565/",
    doi = "10.18653/v1/2022.acl-long.565",
    pages = "8214--8236"
}

@article{jung2022maieutic,
  title={Maieutic Prompting: Logically Consistent Reasoning with Recursive Explanations},
  author={Jung, Jaehun and Qin, Lianhui and Welleck, Sean and Brahman, Faeze and Bhagavatula, Chandra and Bras, Ronan Le and Choi, Yejin},
  journal={arXiv preprint arXiv:2205.11822},
  year={2022}
}

@article{xie2021explanation,
  title={An explanation of in-context learning as implicit bayesian inference},
  author={Xie, Sang Michael and Raghunathan, Aditi and Liang, Percy and Ma, Tengyu},
  journal={arXiv preprint arXiv:2111.02080},
  year={2021}
}

@article{srivastava2022beyond,
  title={Beyond the Imitation Game: Quantifying and extrapolating the capabilities of language models},
  author={Srivastava, Aarohi and Rastogi, Abhinav and Rao, Abhishek and Shoeb, Abu Awal Md and Abid, Abubakar and Fisch, Adam and Brown, Adam R and Santoro, Adam and Gupta, Aditya and Garriga-Alonso, Adri{\`a} and others},
  journal={arXiv preprint arXiv:2206.04615},
  year={2022}
}

@article{dohan2022language,
  title={Language Model Cascades},
  author={Dohan, David and Xu, Winnie and Lewkowycz, Aitor and Austin, Jacob and Bieber, David and Lopes, Raphael Gontijo and Wu, Yuhuai and Michalewski, Henryk and Saurous, Rif A and Sohl-Dickstein, Jascha and others},
  journal={arXiv preprint arXiv:2207.10342},
  year={2022}
}

@article{rae2021scaling,
  title={Scaling language models: Methods, analysis \& insights from training {Gopher}},
  author={Rae, Jack W and Borgeaud, Sebastian and Cai, Trevor and Millican, Katie and Hoffmann, Jordan and Song, Francis and Aslanides, John and Henderson, Sarah and Ring, Roman and Young, Susannah and others},
  journal={arXiv preprint arXiv:2112.11446},
  year={2021}
}

@article{kojima2022large,
  title={Large Language Models are Zero-Shot Reasoners},
  author={Kojima, Takeshi and Gu, Shixiang Shane and Reid, Machel and Matsuo, Yutaka and Iwasawa, Yusuke},
  journal={arXiv preprint arXiv:2205.11916},
  year={2022}
}

@article{gpt3,
 author = {Brown, Tom and Mann, Benjamin and Ryder, Nick and Subbiah, Melanie and Kaplan, Jared D and Dhariwal, Prafulla and Neelakantan, Arvind and Shyam, Pranav and Sastry, Girish and Askell, Amanda and Agarwal, Sandhini and Herbert-Voss, Ariel and Krueger, Gretchen and Henighan, Tom and Child, Rewon and Ramesh, Aditya and Ziegler, Daniel and Wu, Jeffrey and Winter, Clemens and Hesse, Chris and Chen, Mark and Sigler, Eric and Litwin, Mateusz and Gray, Scott and Chess, Benjamin and Clark, Jack and Berner, Christopher and McCandlish, Sam and Radford, Alec and Sutskever, Ilya and Amodei, Dario},
 journal = {Neural Information Processing Systems (NeurIPS)},
 title = {Language Models are Few-Shot Learners},
 year = {2020}
}

@article{wang2022self,
  title={Self-consistency improves chain of thought reasoning in language models},
  author={Wang, Xuezhi and Wei, Jason and Schuurmans, Dale and Le, Quoc and Chi, Ed and Zhou, Denny},
  journal={arXiv preprint arXiv:2203.11171},
  year={2022}
}

@article{zhou2022least,
  title={Least-to-Most Prompting Enables Complex Reasoning in Large Language Models},
  author={Zhou, Denny and Sch{\"a}rli, Nathanael and Hou, Le and Wei, Jason and Scales, Nathan and Wang, Xuezhi and Schuurmans, Dale and Bousquet, Olivier and Le, Quoc and Chi, Ed},
  journal={arXiv preprint arXiv:2205.10625},
  year={2022}
}

@article{liu2021token,
  title={A token-level reference-free hallucination detection benchmark for free-form text generation},
  author={Liu, Tianyu and Zhang, Yizhe and Brockett, Chris and Mao, Yi and Sui, Zhifang and Chen, Weizhu and Dolan, Bill},
  journal={arXiv preprint arXiv:2104.08704},
  year={2021}
}

@article{Nye2021sequence,
 author = {Nye, Maxwell and Tessler, Michael and Tenenbaum, Josh and Lake, Brenden M},
 journal = {Neural Information Processing Systems (NeurIPS)},
 title = {Improving Coherence and Consistency in Neural Sequence Models with Dual-System, Neuro-Symbolic Reasoning},
 year = {2021}
}

@article{wang2022rationale,
  title={Rationale-Augmented Ensembles in Language Models},
  author={Wang, Xuezhi and Wei, Jason and Schuurmans, Dale and Le, Quoc and Chi, Ed and Zhou, Denny},
  journal={arXiv preprint arXiv:2207.00747},
  year={2022}
}

@article{zhao2021calibrate,
  title = 	 {Calibrate Before Use: Improving Few-shot Performance of Language Models},
  author =       {Zhao, Zihao and Wallace, Eric and Feng, Shi and Klein, Dan and Singh, Sameer},
  journal = 	 {International Conference on Machine Learning (ICML)},
  year = 	 {2021},
}

@article{zelikman2022star,
  title={{STaR}: Bootstrapping reasoning with reasoning},
  author={Zelikman, Eric and Wu, Yuhuai and Goodman, Noah D},
  journal={arXiv preprint arXiv:2203.14465},
  year={2022}
}

@article{creswell2022selection,
  title={Selection-Inference: Exploiting Large Language Models for Interpretable Logical Reasoning},
  author={Creswell, Antonia and Shanahan, Murray and Higgins, Irina},
  journal={arXiv preprint arXiv:2205.09712},
  year={2022}
}

@article{nye2021show,
  title={Show your work: Scratchpads for intermediate computation with language models},
  author={Nye, Maxwell and Andreassen, Anders Johan and Gur-Ari, Guy and Michalewski, Henryk and Austin, Jacob and Bieber, David and Dohan, David and Lewkowycz, Aitor and Bosma, Maarten and Luan, David and others},
  journal={arXiv preprint arXiv:2112.00114},
  year={2021}
}

@article{gpt2,
  title={Language Models are Unsupervised Multitask Learners},
  author={Radford, Alec and Wu, Jeff and Child, Rewon and Luan, David and Amodei, Dario and Sutskever, Ilya},
  year={2019}
}

@article{ouyang2022training,
  title={Training language models to follow instructions with human feedback},
  author={Ouyang, Long and Wu, Jeff and Jiang, Xu and Almeida, Diogo and Wainwright, Carroll L and Mishkin, Pamela and Zhang, Chong and Agarwal, Sandhini and Slama, Katarina and Ray, Alex and others},
  journal={arXiv preprint arXiv:2203.02155},
  year={2022}
}

@article{paszke2019pytorch,
  title={Pytorch: An imperative style, high-performance deep learning library},
  author={Paszke, Adam and Gross, Sam and Massa, Francisco and Lerer, Adam and Bradbury, James and Chanan, Gregory and Killeen, Trevor and Lin, Zeming and Gimelshein, Natalia and Antiga, Luca and others},
  journal={Neural Information Processing Systems (NeurIPS)},
  year={2019}
}

@ARTICLE{EM,
  author = {A. P. Dempster and N. M. Laird and D. B. Rubin},
  title = {Maximum likelihood from incomplete data via the {EM} algorithm},
  journal = {Journal of the Royal Statistical Society B},
  year = {1977},
  volume = {39},
  number = {1},
  pages = {1--38}
  }

@article{chowdhery2022palm,
  title={{PaLM}: Scaling language modeling with pathways},
  author={Chowdhery, Aakanksha and Narang, Sharan and Devlin, Jacob and Bosma, Maarten and Mishra, Gaurav and Roberts, Adam and Barham, Paul and Chung, Hyung Won and Sutton, Charles and Gehrmann, Sebastian and others},
  journal={arXiv preprint arXiv:2204.02311},
  year={2022}
}

@article{wei2022chain,
  title={Chain of thought prompting elicits reasoning in large language models},
  author={Wei, Jason and Wang, Xuezhi and Schuurmans, Dale and Bosma, Maarten and Chi, Ed and Le, Quoc and Zhou, Denny},
  journal={arXiv preprint arXiv:2201.11903},
  year={2022}
}

@article{ye2022unreliability,
  title={The Unreliability of Explanations in Few-Shot In-Context Learning},
  author={Ye, Xi and Durrett, Greg},
  journal={arXiv preprint arXiv:2205.03401},
  year={2022}
}

@article{bengio1,
  title={The consciousness prior},
  author={Bengio, Yoshua},
  journal={arXiv preprint arXiv:1709.08568},
  year={2017}
}

@article{bengio2,
  title={Inductive biases for deep learning of human cognition},
  author={Goyal, Anirudh and Bengio, Yoshua},
  journal={arXiv preprint arXiv:2011.15091},
  year={2020}
}

@article{tversky1974judgment,
  title={Judgment under Uncertainty: Heuristics and Biases: Biases in judgments reveal some heuristics of thinking under uncertainty.},
  author={Tversky, Amos and Kahneman, Daniel},
  journal={Science},
  volume={185},
  number={4157},
  pages={1124--1131},
  year={1974},
  publisher={American association for the advancement of science}
}

@book{kahneman,
  title={Thinking, fast and slow},
  author={Kahneman, Daniel},
  year={2011},
  publisher={Macmillan}
}

@article{decomp,
    title={Decomposed Prompting: A Modular Approach for Solving Complex Tasks},
    author={Tushar Khot and Harsh Trivedi and Matthew Finlayson and Yao Fu and Kyle Richardson and Peter Clark and Ashish Sabharwal},
    journal={arXiv preprint arXiv:2210.02406},
    year={2022}
}

@article{suzgun2022challenging,
  title={Challenging BIG-Bench tasks and whether chain-of-thought can solve them},
  author={Suzgun, Mirac and Scales, Nathan and Sch{\"a}rli, Nathanael and Gehrmann, Sebastian and Tay, Yi and Chung, Hyung Won and Chowdhery, Aakanksha and Le, Quoc V and Chi, Ed H and Zhou, Denny and others},
  journal={arXiv preprint arXiv:2210.09261},
  year={2022}
}

@article{ape,
    title={Large Language Models Are Human-Level Prompt Engineers},
    author={Yongchao Zhou and Andrei Ioan Muresanu and Ziwen Han and Keiran Paster and Silviu Pitis and Harris Chan and Jimmy Ba},
    journal={arXiv preprint arXiv:2211.01910},
    year={2022}
}

@article{chung2022scaling,
  title={Scaling instruction-finetuned language models},
  author={Chung, Hyung Won and Hou, Le and Longpre, Shayne and Zoph, Barret and Tay, Yi and Fedus, William and Li, Eric and Wang, Xuezhi and Dehghani, Mostafa and Brahma, Siddhartha and others},
  journal={arXiv preprint arXiv:2210.11416},
  year={2022}
}
\bibliographystyle{acl_style/acl_natbib}

\counterwithin{figure}{section}
\counterwithin{table}{section}

\appendix
\onecolumn

\section{Additional tasks}
\label{sec:add_exp}

Descriptions of all the tasks studied here can be found in \S\ref{sec:app-tasks}.

\subsection{Uncertainty and hallucination detection}
\label{sec:exp_uncertainty}

LLMs are prone to generating hallucinations that contain incorrect statements. The likelihoods of these statements are often dominated by short plausible patterns, which also makes it difficult for LLMs to evaluate their own uncertainty about a fact. Thus, detection \citep{liu2021token,zhou-etal-2021-detecting} and reduction of such hallucinations is crucial for widespread use of LLMs in real applications \citep{dziri-etal-2021-neural,shuster-etal-2021-retrieval-augmentation}.

\subsubsection{Sports understanding}

\begin{figure*}[ht]
    \centering
   \includegraphics[width=0.3\linewidth]{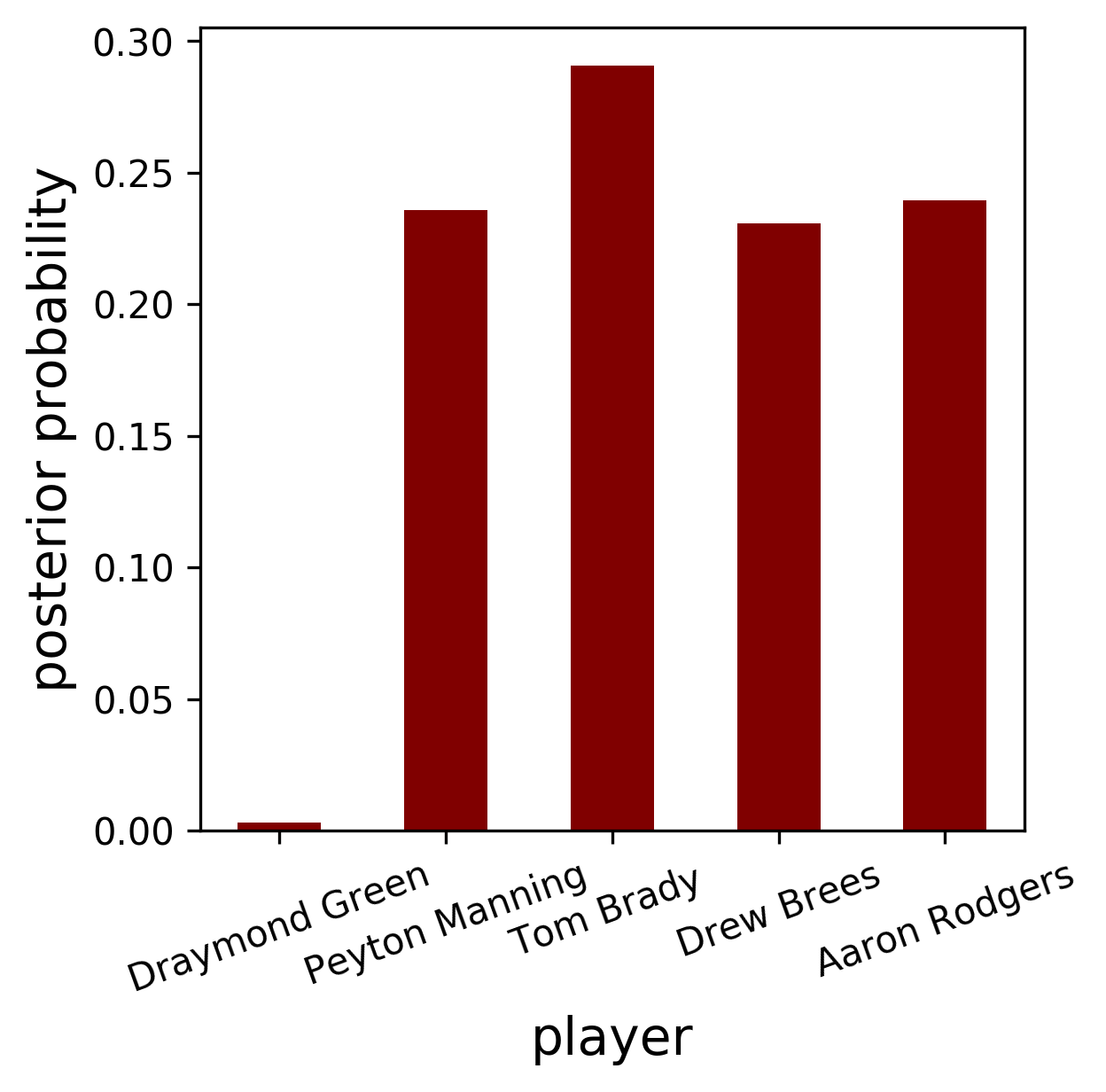}
   \caption{Example posterior probabilities generated from text-davinci-002 for \taskname{Sports understanding} with the description \textit{``threw a touchdown''}.  The basketball player given in the question \textit{Draymond Green} has a much lower posterior probability than the generated football players, from which we conclude the sentence \textit{``Draymond Green threw a touchdown.''} is implausible.}
    \label{fig:sports}
\end{figure*}

Questions in \textsc{Sports understanding} ask to determine whether it is `plausible' or `implausible' that a professional sports player $x$ (e.g., `Draymond Green', a basketball player) performed an action $a$ associated with a sport (e.g., `threw a touchdown', an action in American football). It is implied that the combination of $x$ and $a$ is plausible if the sport with which player $x$  is associated coincides with the sport in which action $a$ is performed. We consider an approach that does not rely on identifying the latent variable (sport) as an intermediate step and is thus more generalizable to other domains.

We use an \thinktechnique{Example generation} \Think{} prompt to produce a set $S$ of players who perform action $a$, then do \sumtechnique{Posterior computation} by normalizing the likelihood assigned by the LLM to each player in $S$, as well as $x$, performing action $a$:
\[\forall y\in S\cup\{x\}\quad p(y|a)=\frac{p_{\LM}(\text{``$y$ $a$''})}{\sum_{y'\in S\cup\{x\}}p_{\LM}(\text{``$y'$ $a$''})}\]
The statement is considered to be implausible if the posterior on $x$ is sufficiently low (\sumtechnique{Thresholding}) -- see Fig.~\ref{fig:sports}.

\subsubsection{Known unknowns}

Questions in the \taskname{Known unknowns} task ask to determine whether the answer to a question is a certain precise concept or `unknown'. %

Given a question $q$ (e.g., ``What was the temperature in Cuzco on the day of the Emperor Vespasian's birth'') and the candidate precise answer $a$ (e.g., $25^\circ$C), we use a \thinktechnique{List extension} prompt to generate a set $S$ of other possible answers to $q$. We then do a \sumtechnique{Posterior computation} over $S$ and the original answer $a$, similar to that used for \taskname{Sports understanding}:
\[\forall y\in S\cup\{a\}\quad p(y|q)=\frac{p_{\LM}(\text{``\texttt{$q$? $y$}''})}{\sum_{y'\in S\cup\{a\}}p_{\LM}(\text{``\texttt{$q$? $y'$}''})}.\]
The answer $a$ is chosen  if the posterior on $a$ is sufficiently high (\sumtechnique{Thresholding}), and otherwise `unknown' is chosen.

\subsection{Translation between languages and writing systems}
\label{sec:exp_translation}

This extends the results on \taskname{Logical deduction} in \S\ref{sec:exp_logical}.

\subsubsection{Russian misconceptions.} 

In the \taskname{Misconceptions Russian} task, the true statement must be chosen out of a pair of Russian sentences: a statement $s$ and its negation $t$.

We first describe an approach that does not use translation and already performs better than random guessing -- and better than baseline methods that simply select the more likely of the two statements -- using the largest GPT-3 model, which has sufficient knowledge of Russian. 
We compute the posterior over the two hypotheses ``$s$ is true, $t$ is false'' and ``$s$ is false, $t$ is true'':
\[\small
p_{\LM}(\text{``\texttt{T}''}\mid\text{``\texttt{T or F? $s$. Answer: }''})
p_{\LM}(\text{``\texttt{F}''}\mid\text{``\texttt{T or F? $t$. Answer: }''}),
\]
\[\small
p_{\LM}(\text{``\texttt{F}''}\mid\text{``\texttt{T or F? $s$. Answer: }''})
p_{\LM}(\text{``\texttt{T}''}\mid\text{``\texttt{T or F? $t$. Answer: }''}).
\]
where \texttt{T} denotes \texttt{True} and \texttt{F} \texttt{False} in the actual prompt. This is a kind of \sumtechnique{Product aggregation}. If the posterior on the first option is higher, $s$ is chosen as the true statement; otherwise, $t$ is chosen.

This approach can be combined with a \thinktechnique{Translation} prompt that produces translations of $s$ and $t$ into English, then uses these translations in place of $s$ and $t$ in the above computations. The approach can be further extended by sampling a \emph{set} of translations and performing \sumtechnique{Mixture aggregation} over the translations. Our reported result uses 10 generated translation for each statement, but it is only $2\%$ higher than the result using one generated translation.

\subsubsection{Emoji movie} 

The multiple-choice \taskname{Emoji movie} task requires selecting the name of a movie from a list $\{m_i\}$ that is best described by a sequence of emoji symbols $s=(s_1\dots s_n)$. An \thinktechnique{Order inversion} prompt performs best on this task using the Davinci variant of GPT-3: choosing the answer
\[\argmax_ip_{\LM}(s\mid\text{``\texttt{Emoji describing the movie $m_i$}''}).\]
We also attempt to use a \thinktechnique{Translation} prompt to obtain a single-word English description $w_j$ of each emoji $s_j$ in $s$, then score using 
\[\argmax_ip_{\LM}(w_1\dots w_n\mid\text{``\texttt{Words describing the movie $m_i$}''}).\]
This approach performs slightly better than \thinktechnique{Order inversion} alone using InstructGPT. However, it does not work with the base GPT-3 models, which do not as reliably translate emoji to English.

\subsubsection{Persian QA}

We solve this standard extractive question answering task by simply translating the passage and question from Persian to English using a \textbf{Translation} prompt, generating English text, up to the first period or line break, following the concatenation of the translated prompt and question, and translating the result back to Persian using another \textbf{Translation} prompt.

No few-shot algorithms have above zero accuracy on this task, indicating models' knowledge is sufficient to translate between languages (probably due to the presence of paired data in the training corpus), but insufficient to reason in the source language without passing through an intermediate latent variable, the translation.

Finally, note that the accuracy is evaluated by exact string match, which contributes to the very low scores. We observed that the answers generated by \ThinkSum{} are often paraphrases or terms related to the correct answers, which suggests that the result could be improved by using the knowledge that the target string always appears verbatim as a substring of the prompt.

\subsection{Semantic relatedness}
\label{sec:exp_relatedness}

This extends the results on \taskname{Odd one out} in \S\ref{sec:exp_ooo}.

\subsubsection{Phrase relatedness} 

Each question in the multiple-choice \taskname{Phrase relatedness} task requires to determine which of a given set of words or phrases $\{w_i\}$ is related to a query phrase $q$. We query the LLM for the likelihood of $q$ following a \thinktechnique{List of words} prompt to form a vector of likelihoods:
\begin{equation*}
p_i=p_{\LM}(q\mid\text{``\texttt{List of words: $w_i$, }''}).
\end{equation*}
The answer selected is the one with highest likelihood, $\argmax_ip_i$ (a trivial \Sum{} operation).
We note that this is also an instance of \thinktechnique{Order inversion}: the query is scored following a prompt in which each of the candidate answers is substituted.

\subsubsection{Codenames} 

Each question in \taskname{Codenames} requires selecting the $k$ words from a set $\{w_i\}$ that are most closely related to a query word $q$. We form a vector $p_i$ in the same way as for \taskname{Phrase relatedness}, then select the top $k$ entries in $p_i$ to produce the output.\footnote{Because the task is evaluated by BLEU score against the reference answers listed in alphabetical order, we perform the additional step of converting the top indices to the answer in the right format. Alphabetization of short lists is trivial in code, but can also very reliably be done by prompting GPT-3.}

\subsection{Substitution and aggregation}
\label{sec:exp_substitution}

We give two other example of substitution and aggregation operations complementing the experiments on \taskname{Invented words} (\S\ref{sec:exp_binnebam}) and \taskname{Odd one out} (\S\ref{sec:exp_ooo}).

\subsubsection{Novel concepts}

In the multiple-choice \taskname{Novel concepts} task, a set of words or phrases $W=\{w_i\}$ and a set of statements $S=\{s_j\}$ with third-person plural pronoun subjects (`They all...') are given, and the statement which is true for all items in $W$ must be determined. 

We treat each statement $s_j$ as a \emph{template}, into which words $w$ can be substituted by replacing `They all' with $w$. Denoting by $s_j\langle w\rangle$ the substitution of $w$ into $s_j$, we form a $|W|\times|S|$ matrix $P_{ij}$ by scoring the \thinktechnique{Substitution} of each word into each statement and considering the \sumtechnique{Ratio of likelihoods} with the template without substitution:
$
    P_{ij}=\frac{p_{\LM}(s_j\langle w_i\rangle)}{p_{\LM}(s_j)}.
$
We then perform \sumtechnique{Product aggregation} to select the statement which is most likely to be generated by all words in the set. To be precise, the selected statement is $\argmax_j\prod_iP_{ij}$.

\subsubsection{Code line description}

We solve the \taskname{Code line description} task, in which a correct comment for a code snippet is to be chosen, using \thinktechnique{Order inversion} and \thinktechnique{Substitution} techniques.

The greatest gain -- amounting for all but 1\% of the improvement relative to direct prompting -- arises from \thinktechnique{Order inversion}. Instead of ranking the candidate comments $c$ by their likelihood following the given code $s$ (i.e., $p(c|s)$), we score each candidate comment $c$ by the likelihood of the code to follow $c$ formatted as a Python comment ($p(s|$``\texttt{\# }$c$'')).

We also experimented with \thinktechnique{Substitution} and \sumtechnique{Product aggregation}, which yielded an additional small accuracy gain. The code snippets are written in Python, which requires code to be formatted using an arbitrary but consistent number of spaces for line indentation. Using the knowledge that the correct comment should be most likely to generate the program in \textit{any} of its equivalent representations, we scored comments in the manner described in the preceding paragraph, but with $s$ reformatted with different number of indentation spaces $n$. The resulting scores were then multiplied over $n=1,2,\dots,6$ and the highest-scoring comment selected.

\subsection{Other tasks}
\label{sec:exp_other}

\subsubsection{Language identification}

The multiple choice \taskname{Language identification} task is similar in form and solution to \taskname{Code line description} and we include it for completeness to show the large difference that can be made by \thinktechnique{Order inversion}.

Rather than scoring all candidate language names $\ell$ following the given sentence $s$ (i.e., $p(s|\ell)$), we instead score each language name $\ell$ by $p(s|``$\texttt{The following is a sentence in }$\ell$\texttt{:}''$)$ and select the highest-scoring $\ell$ as the answer.

\begin{table}[t]
\centering
\small
\begin{tabular}{ll}
\toprule
\taskname{Odd one out} & 
\begin{minipage}{0.7\textwidth}
Words: blue, pink, magenta, banana\\
All words are colors except banana. The odd one out is banana.\\
 \\
Words: pencil, eraser, baby, rule, notebook\\
All words are office supplies except baby. The odd one out is baby.
\end{minipage}
\\\midrule
\taskname{Phrase relatedness} & 
\begin{minipage}{0.7\textwidth}
For each word or phrase, identify the most related choice from the listed options.\\
Input: Ice Cream\\
Option: Antarctica\\
Option: Titanic\\
Option: Dessert\\
Option: Sour Cream\\
Ice cream is a type of dessert. Therefore, ice cream and dessert are the most related.\\
Answer: Dessert
\end{minipage}
\\\midrule
\taskname{Known unknowns} & 
\begin{minipage}{0.7\textwidth}
What was the population of San Francisco in 2018?\\
Option: 879,676\\
Option: Unknown\\
The question asks the population of San Francisco in 2018, for which data can be collected. Population data for cities on a yearly basis is available, and thus the answer is known, and it is 879,676. \\
Answer: 879,676\\
What was the population of San Francisco yesterday?\\
Option: 891,402\\
Option: Unknown\\
The question asks the population of San Francisco yesterday. As it is not possible to know the exact population of a city on a daily basis, the answer for this question is unknown.\\
Answer: Unknown
\end{minipage}
\\\midrule
\taskname{Logical deduction} & 
\begin{minipage}{0.7\textwidth}
On a table, there are five plates: a black plate, a white plate, a green plate, a blue plate, and a red plate. The white plate is bigger than the green plate.  The red plate is the biggest. The black plate is bigger than the blue plate. The black plate is smaller than the green plate. Which plate is the smallest?\\
Option: The red plate is the smallest.\\
Option: The black plate is the smallest.\\
Option: The white plate is the smallest.\\
Option: The green plate is the smallest.\\
Option: The blue plate is the smallest.\\
The black plate is bigger than the blue plate. The black plate is smaller than the green plate, as a result the green plate is bigger than the blue plate as well. The white plate is bigger than the green plate, which is bigger than the blue plate. As a result, the green plate is bigger than the blue plate. The red plate is the biggest, so it is bigger than the blue plate. Since all other plates are bigger than the blue plate, the blue plate is smallest.\\
Answer: The blue plate is the smallest.
\end{minipage}
\\\midrule
\taskname{Invented words} & 
\begin{minipage}{0.7\textwidth}
The word 'borger' are animals who bite specific things for fun, and the word 'folpt' is a type of a chewy toy. Question: Which of the following sentences best characterizes borger folpts?\\
Option: Borger folpts are leashes for animals.\\
Option: Borger folpts are toys for infants.\\
Option: Borger folpts are hard to swallow.\\
Option: Borger folpts are pet toys.\\
Borgers are animals, and folpts are chewy toys. Therefore, borger folpts are chewy toys that animals, or pets, can play with. Therefore, the answer is borger folpts are pet toys.\\
Answer: Borger folpts are pet toys.
\end{minipage}
\\\bottomrule
\end{tabular}
\caption{Few-shot demonstrations used for chain of thought (Table~\ref{tab:cot_results}).}
\label{tab:cot_prompt_examples}
\end{table}
\section{BIG-bench Lite}

\begin{figure}[t]
    \centering
    \includegraphics[width=0.7\textwidth]{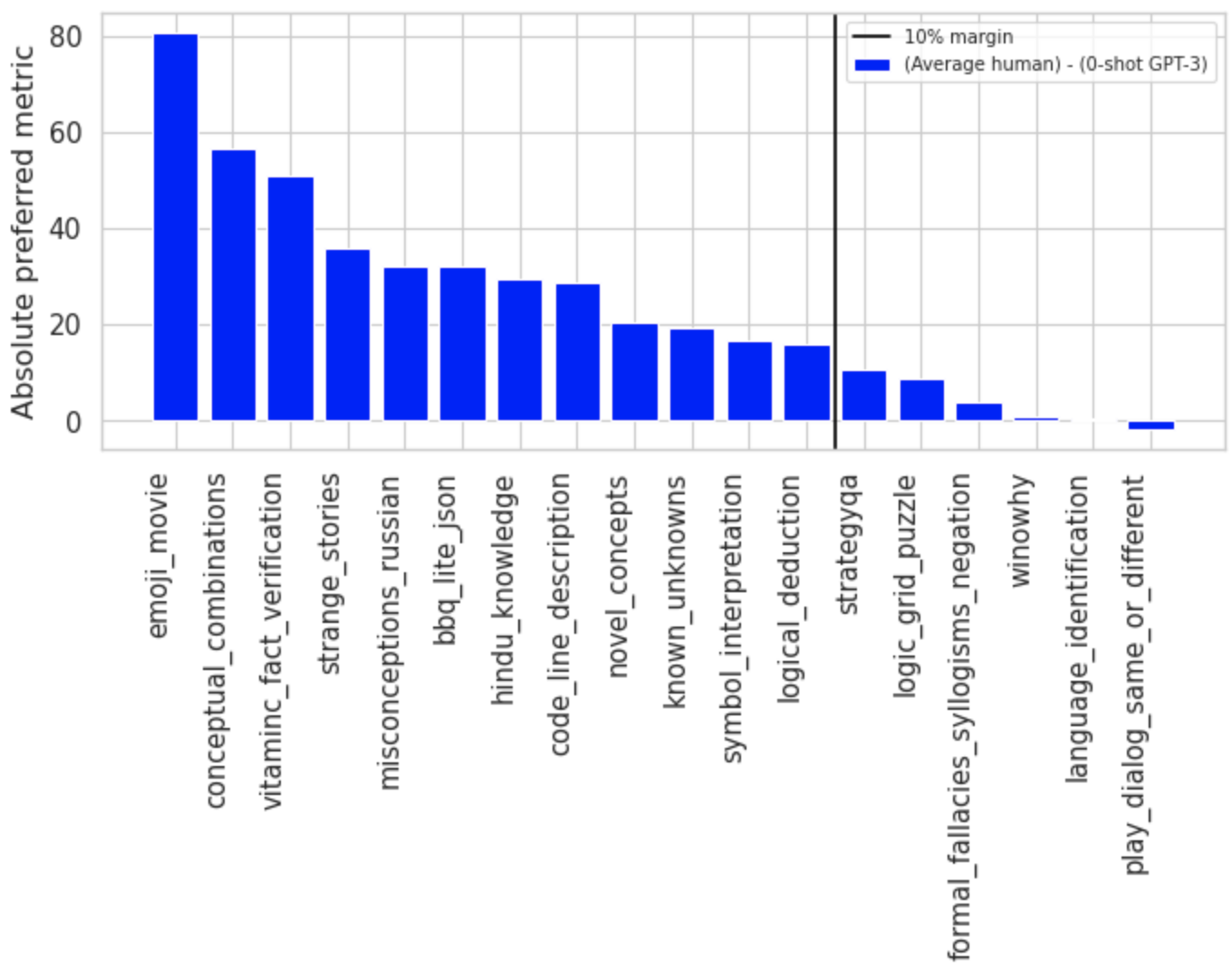}
    \caption{Margin between 0-shot GPT-3 and average human performance for BIG-bench Lite tasks. Using \ThinkSum{}, we address many of the tasks that have greater than $10\%$ performance margin with average human, and significantly reduce and often overturn the margin.}
    \label{fig:margins}
\end{figure}

Figure \ref{fig:margins} shows the performance margin between an average human and zero-shot GPT-3 on tasks in BIG-bench Lite, a select subset of tasks chosen by the authors of the benchmark to showcase the most important aspects of LLMs that need improvement. The vertical black bar separates the dataset into tasks where GPT-3 is already within the margin of just 10\% compared to the average human accuracy, and the harder tasks (on the left).  We show in the main text that some of these harder tasks, in particular \taskname{Emoji movie}, \taskname{Conceptual combinations},\taskname{Known unknowns}, \taskname{Novel concepts}, \taskname{Misconceptions Russian} and \taskname{Logical deduction}, the margins are shrunk considerably, often exceeding average human performance. Other tasks in BIG-bench lite such as \taskname{Logic grid puzzle} and  \taskname{Symbol interpretation} share a similar structure to the addressed by \ThinkSum{}, and thus could be investigated as part of future work. Another example where \ThinkSum{} can be applied is the \taskname{Code line description} task, where we observe in our preliminary experiments that a simple order inversion can significantly outperform average human accuracy. 

\section{Task descriptions}
\label{sec:app-tasks}

\subsection{Hallucination detection}
\paragraph{Known unknowns.} In this task, the aim is to measure the ability of LLMs to identify whether the answer to a question is known, or unknown. If the answer to the question cannot be known, an imaginary answer is generated, and the model is asked to choose between the imaginary answer and the word ``Unknown''. If the question has a correct answer, the correct answer and ``Unknown" are provided as options. An example question from this task is \textit{``How often did Abraham Lincoln cut his toenails?''}, and the options are \textit{``Every Saturday night'',``Unknown''}. Similar to the sports understanding task, we can use an LLM to generate other plausible answers with the following prompt: \textit{How often did Abraham Lincoln cut his toenails? A possible answer is: \textbackslash n 1. Every Saturday night. List 4 other possible answers in the same format as the first: \textbackslash n 2.} Then, the answer given in the question is predicted to be ``known'' if its posterior is higher by the second most likely option by some margin. In our experiments, we chose this value to be $1/N_e$ where $N_e$ is the number of examples, including the original option. 

\paragraph{Sports understanding.} 
In this task, artificial sentences are constructed using the name of professional sports players and actions from particular sports. The model is then asked to identify whether the sentence is plausible, where a sentence is considered plausible if the sport of the player matches the sport of the action described in the sentence. An example from the task is \textit{``Statement: Draymond Green threw a touchdown. Plausible/implausible?''}

For \ThinkSum{} to be able to parse outputs, GPT-3 generations need to be in a pre-determined format. While larger models can obey a format without demonstrations, for smaller models it is helpful to demonstrate the format with an example. Thus, we use the following prompt: ``\textit{List 4 examples of players who scored a rabona goal.\textbackslash n 1. Cristiano Ronaldo\textbackslash n 2. Erik Lamela\textbackslash n 3. Mario Balotelli\textbackslash n 4. Angel Di Maria\textbackslash n List 4 examples of players who threw a touchdown.\textbackslash n1.''}.

\subsection{Translation}

\paragraph{Emoji movie.} In this task, the LLM is presented with emojis that describe the plot of a movie, and the model is asked to guess the correct movie from the given options. The task evaluates the LLMs understanding of emojis, and its knowledge of popular movies. An example from the task is \textit{``What movie does this emoji describe? \{lying-face-emoji\} \textbackslash n, pinocchio \textbackslash n heat \textbackslash n the big sleep, \textbackslash n inception \textbackslash n lilo \& stitch}

\paragraph{Misconceptions Russian.} In this task, the LLM is presented with sentences in Russian that contain misconceptions, and their true counterparts. The task evaluates if the LLM believes common misconceptions. One example of a target pair from the task, translated to English is as follows: \textit{``Vaccines do not cause autism. \textbackslash n Vaccines cause autism.''}

\paragraph{ParsiNLU reading comprehension.} This is a standard extractive QA task in which a short passage and question are provided in Persian and the model must generate the answer, also in Persian.

\subsection{Semantic relatedness}
\paragraph{Phrase relatedness.} In this task, an input phrase, and four target phrases are presented to the language model. The language model is asked to identify the most related choice from the listed target options. An example from the task is \textit{``For each word or phrase, identify the most related choice from the listed options. \textbackslash n Input: home town \textbackslash n Option: town center  \textbackslash n Option: location \textbackslash n Option: native city \textbackslash n Option: home run''}

\paragraph{Codenames.} In this task, the language model is asked to identify words associated with a given word. 
An example from the task is \textit{``Try to identify the 2 words best associated with the word WHITE from the following list: \textbackslash n book, anchor, rainbow, shoulder, tunnel, sack, drum, pacific, page, mark, gear, glacier. Give your answer in alphabetical order.''} 

\paragraph{Odd one out.} This task is aimed at evaluating the capability of LLMs in semantic relatedness. This task presents the model with four to six words, where all words except one word are semantically or grammatically related to each other. The goal for the language model is to identify the odd word. An example question from the task is \textit{``Pick the odd word out: glass, head, arm, leg, hand, foot''}.

\subsection{Concept understanding}

In the following tasks, the shared goal is to test the ability of LLMs on concepts over entities that have likely not been observed during training.

\paragraph{Conceptual combinations: Invented words.} 
In this task, the LLM is provided with two invented words, and their definitions in the input. The LLM is then asked to infer the most plausible meaning resulting from the combination of the invented words. As the words are invented, they are not present in the training set, and the LLM needs to understand and combine the definitions of the invented words to reason about the meaning of the combination. An example is: \textit{``The word 'binne' means any animal that is furry and has four legs, and the word 'bam' means a simple sort of dwelling. Question: Which of the following sentences best characterizes binne bams?''}. Similar to \taskname{Sports understanding}, we can use the following prompt to force the LLM to obey a fixed format: \textit{``List synonyms of binne, separate synonyms by comma:''}

\paragraph{Novel concepts.} In this task, the LLM is presented with two to four disparate entities that typically would not co-occur frequently, but share an underlying conceptual or linguistic concept. The aim is to test the ability of the LLM to reason about entities that are unlikely to have been observed in the same context during training. In a multiple-choice setting, the LLM is given concepts relating to the entities, and is asked to generate the intended concepts against carefully chosen tempting distractors. The choices are not presented in the prompt. An example question from the task is as follows: \textit{``What do the following have in common? 1) bumble bees 2) 01010101 3) race cars''}, and the answer options are \textit{They all make noise,} \textit{``They all are yellow,} \textit{They all are binary,} \textit{They all go fast,} \textit{They all have stripes''}.

\subsection{Other tasks}

Two multiple-choice tasks test the LLM's knowledge of specific domains, such as uncommon languages and programs.

\paragraph{Code line description.}

This task requires the LLM to select the appropriate text description, out of four choices, for a short snippet of Python code, that could act as a comment describing the behaviour of a function.

\subsubsection{Language identification.}

This task requires the LLM to select, out of eleven choices, the language in which a text is written. The languages represent a diversity of language families and writing systems and most are very infrequent in text found on the Internet.

\section{Additional experimental details}

Our experiments are performed using four different sizes of GPT-2 (Small, Medium, Large, and XL) \citep{gpt2}, GPT-3 with four different model sizes (ada,babbage,curie,davinci) \citep{gpt3}, and InstructGPT \citep{ouyang2022training}. All GPT-3 experiments are run between August 2022 and September 2022 by using the OpenAI API. Our GPT-2 experiments were run in PyTorch \citep{paszke2019pytorch} and the Hugging Face Transformers library with a Tesla K80 GPU.

\subsection{Hyperparameters}

\paragraph{Maximum generation length.} For tasks that require \thinktechnique{example and list generation}, such as \taskname{conceptual combinations}, \taskname{Known unknowns}, and \taskname{Sports understanding}, we use $\text{max}\_\text{tokens} = 100$. For \thinktechnique{fact generation} in \taskname{Odd one out} with auxiliary knowledge and \ThinkSum{}, we use $\text{max}\_\text{tokens} = 1000$.

\paragraph{Temperature.} All GPT-2 experiments used $\text{temperature} = 0.5$. For \taskname{Sports understanding} and translation tasks, we used $\text{temperature} = 0.5$ to promote diversity of generated plausible options. All other experiments used $\text{temperature} = 0$ (greedy decoding).

\paragraph{Number of examples ($N_e$).} For \taskname{Conceptual combinations} we used $N_e = 2$, and for \taskname{Known unknowns} and \taskname{Sports understanding} we used $N_e = 4$.

\paragraph{Threshold.} A threshold of $0.01$ was used for \taskname{Sports understanding}.

\subsection{Using an LLM to evaluate inequalities.}
\label{sec:llm_ineq}

\begin{figure}[t]
  \centering
      \includegraphics[width = 0.9\textwidth]{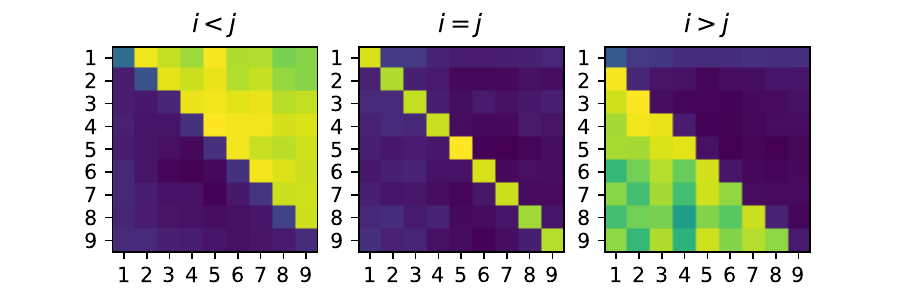}
  \caption{Probabilities of different (in)equalities according to GPT-3 text-davinci-002 (logit).}
 \label{fig:inequality_eval}
\end{figure}

\paragraph{Using GPT-3 or external algorithms to evaluate inequalities.} We show how a LLM can be used to find the truth values of inequalities involving small numbers, rather than resorting to calls to an external system that is aware of arithmetic. Fig.~\ref{fig:inequality_eval} shows the matrix of posterior probabilities evaluated using InstructGPT (text-davinci-002) for strings of form ``$x$\texttt{=}$y$'', ``$x$\texttt{<}$y$'', ``$x$\texttt{>}$y$'' for $x,y \in \{1,..,9\}$. The probabilities are computed using prompts of the form ``\texttt{True or false: }$x$\texttt{<}$y$\texttt{? The answer is:}'' and normalizing the probability of the first token over the two options ``\texttt{true}'' and ``\texttt{false}''. These are the probabilities evaluated in (\ref{eq:ineq_sum}). 

\subsection{Knowledge generation details}
\label{sec:know}
\paragraph{Post-processing.} In our knowledge generation experiments for both \ThinkSum{} and the auxiliary knowledge approach, we post-process the generated knowledge statements, to ensure formatting does not harm the predictions of each method. We first remove the extra spaces and the numbers and punctuation generated by the LLM before each fact while enumerating the items of the list. Later, we only keep sentences that contain only one of the objects of interest from the task, to make sure each sentence contains a knowledge statement into which any of the objects can be substituted. Finally, sentences with less than 3 words are removed as these are not likely to contain informative statements.

\begin{figure}[t]
    \centering
    \includegraphics[width=0.5\textwidth]{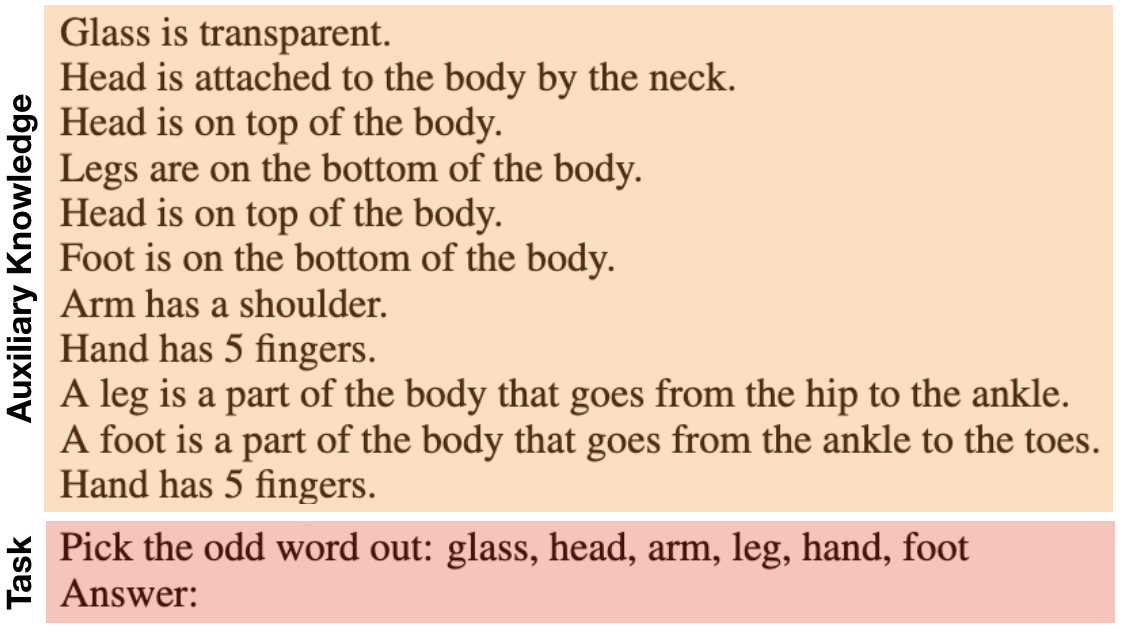}
    \caption{Auxiliary knowledge prompting applied to \taskname{Odd one out}. Facts are generated using the ``list differences'' prompt described in Figure \ref{fig:odd_combined} (right) and post-processed according to \S\ref{sec:know}.}
    \label{fig:aux}
\end{figure}

\paragraph{Auxiliary knowledge.} For auxiliary knowledge experiments, we prepend the generated and post-processed knowledge statements before the question in the task. An example is illustrated in Figure \ref{fig:aux}.

\begin{table}[t]
\centering
\small
\resizebox{\columnwidth}{!}{
\begin{tabular}{p{\textwidth}}
\toprule
\textbf{Task:}  \taskname{Phrase relatedness}  \\
\textbf{Input:} For each word or phrase, identify the most related choice from the listed options. \\
Input: home town \\
Option: town center \\
Option: location \\ 
Option: native city \\ 
Option: home run \\

\midrule

\textbf{Task:} \taskname{Odd one out} \\
\textbf{Input:} Pick the odd word out: glass, head, arm, leg, hand, foot\\

\midrule

\textbf{Task:}  \taskname{Conceptual combinations: Invented words}  \\
\textbf{Input:} The word `binne' means any animal that is furry and has four legs, and the word `bam' means a simple sort of dwelling. Question: Which of the following sentences best characterizes binne bams? \\
Option: Binne bams are for pets.\\
Option: Binne bams are typically furnished with sofas and televisions.\\
Option: Binne bams are luxurious apartments.\\
Option: Binne bams are places where people live\\

\midrule
\textbf{Task:}  \taskname{Novel concepts}  \\
\textbf{Input:} What do the following have in common? 1) bumble bees 2) 01010101 3) race cars \\
Option: They all make noise. \\
Option: They all are yellow. \\
Option: They all are binary. \\
Option: They all go fast. \\
Option: They all have stripes. \\

\midrule
\textbf{Task:}  \taskname{Logical deduction: Five objects}  \\
\textbf{Input:} On a shelf, there are five books: a gray book, a red book, a purple book, a blue book, and a black book. The red book is to the right of the gray book. The black book is to the left of the blue book. The blue book is to the left of the gray book. The purple book is the second from the right. \\
Option: The $\{$gray, red, purple, blue, black$\}$ book is second from the right.\\

\midrule
\textbf{Task:}  \taskname{Codenames} \\
\textbf{Input:} Try to identify the 2 words best associated with the word WHITE from the following list: book, anchor, rainbow, shoulder, tunnel, sack, drum, pacific, page, mark, gear, glacier. Give your answer in alphabetical order. \\

\midrule

\textbf{Task:}  \taskname{Known unknowns} \\
\textbf{Input:} How often did Abraham Lincoln cut his toenails? \\
Option: Every Saturday night\\
Option: Unknown\\

\midrule
\textbf{Task:}  \taskname{Sports understanding} \\
\textbf{Input:} Statement: Draymond Green threw a touchdown. 
Plausible/implausible?\\

\midrule
\textbf{Task:}  \taskname{Emoji movie} \\
\textbf{Input:} What movie does this emoji describe? $\{$lying-face-emoji$\}$ \\
Option: pinocchio\\
Option: heat\\
Option: the big sleep\\
Option: inception\\ 
Option: lilo \& stitch\\
\midrule
\textbf{Task:}  \taskname{Misconceptions Russian} \\
\textbf{Input:} Vaccines cause autism. / Vaccines do not cause autism. \textit{ [in Russian]}\\

\midrule
\textbf{Task:}  \taskname{Code line description} \\
\textbf{Input:}\\
\tt for i in range(23):\\   
\quad\tt print(i)\\
Option: prints values from 0 to 22, \\
Option: computes first 10 prime numbers,\\
Option: prints values from 1 to 10,\\
Option: prints 'hello world' to the terminal
      \\

\midrule
\textbf{Task:}  \taskname{ParsiNLU reading comprehension} \\
\textbf{Input:} To reduce fever, use over-the-counter medications such as acetaminophen and ibuprofen. Note the appropriate dosage and do not use them alongside other fever-reducing medications. You should not give aspirin to your baby without consulting a doctor. Babies under 6 months of age should not be given ibuprofen.\\
What brings down fever? \\
\textit{[in Persian]}\\

\midrule
\textbf{Task:}  \taskname{Language identification} \\
\textbf{Input:} 
Given a sentence, select the correct language among the choices.\\
Mi texaas o a mu vipin simi ri xavil ina vipin si Krais xa. E mi lamon o ne taa siak a xavil ina vipin si Krais e faxuvule xuvul pana vipin sina tefin aava lisan xolane, piau paaliu!\\
Options:
Assamese, 
Nandi, 
Patamona, 
Chavacano, 
Kapingamarangi, 
Turkish, 
Kara, 
Bribri, 
Gofa, 
Pali, 
Shatt  \\

\bottomrule
\end{tabular}}
\caption{List of examples for the studied BIG-bench tasks.}
\label{tab:task_examples}
\end{table}

\subsection{Inference Cost for ThinkSum}

The inference cost for ThinkSum scales with the number of parallel calls to the LLM, which is determined for each task by the number of \Think{} prompts used and the number of objects for which likelihood computations are required at the \Sum{} stage. For the tasks that we considered, as the number of \Think{} prompts is not typically high and the prompts are short, the inference cost increase is marginal. In some cases, \ThinkSum{} is faster than chain-of-thought prompting due to its ability to perform parallel calls to the LLM. For instance, \ThinkSum{} is 23\% faster for \taskname{Phrase Relatedness} compared to chain-of-thought approaches with 5 facts generated using InstructGPT.

\section{Expectation Maximization}
\label{sec:em}

We model items $i \in I$ and facts $f \in F$ as being generated from a latent class $c \in \{0,1\}$. The  distribution is modeled as:
\begin{align*}
    P(i,f \mid  c) = P(i \mid  c)P(f \mid  c) \quad P(i,f)=\sum_cP(c)P(i,f \mid  c)
\end{align*}
where $P(i,f)$ is a matrix of likelihoods from the LLM and the semantic components, groupings $P(i \mid  c)$ and $P(f \mid  c)$. The iterative expectation-maximization \cite[EM;][]{EM} algorithm to derive $P(i \mid  c)$ and $P(f \mid  c)$ has the following updates:
\begin{align*}
    Q(c \mid  i,f) &\propto P(i \mid  c)P(f \mid  c)P(c) \\
    P(i \mid  c) &\propto \sum_f P(i,f)Q(c \mid  i,f) \\
    P(f \mid  c) &\propto \sum_i P(i,f)Q(c \mid  i,f) \\
    P(c) &\propto \sum_{i,f} P(i,f)Q(c \mid  i,f) 
\end{align*}
where $Q(c\mid i,f)$ is the posterior distribution over the latent class $c$ that we maintain for each pair $(i,f)$. EM is run for $200$ iterations, which is more than sufficient for convergence.

\end{document}